\documentclass[10pt,twocolumn,letterpaper]{article}

\usepackage{cvpr}

\usepackage[T1]{fontenc}    % use 8-bit T1 fonts
\usepackage{url}            % simple URL typesetting
\usepackage{booktabs}       % professional-quality tables
\usepackage{amsfonts}       % blackboard math symbols
\usepackage{nicefrac}       % compact symbols for 1/2, etc.
\usepackage{microtype}      % microtypography

\usepackage[utf8]{inputenc}
\usepackage{graphicx}
\usepackage{amsmath}
\usepackage{bm}
\usepackage{subcaption}
\usepackage{placeins}
\usepackage{wrapfig}
\usepackage[font=small,labelfont=bf]{caption}
\usepackage[english]{babel}
\usepackage{sidecap}
\usepackage{enumitem}
\usepackage{xcolor}
\usepackage{nidanfloat}
\usepackage[numbers]{natbib}

\usepackage[pagebackref=true,breaklinks=true,colorlinks,bookmarks=false]{hyperref}

\newcommand{\bw}{{\ensuremath{\bm{w}}}}

\cvprfinalcopy % *** Uncomment this line for the final submission

\def\cvprPaperID{6159} % *** Enter the CVPR Paper ID here

\ifcvprfinal\fi
\begin{document}

\title{Sequential mastery of multiple visual tasks:\\Networks naturally learn to learn and forget to forget}
%metalearning and catastrophic forgetting scale}

%\icmlkeywords{metalearning,catastrophic forgetting, multitask learning, scaling}

\author{%
  Guy Davidson\\
  %\thanks{Use footnote for providing further information about author (webpage, alternative address)---\emph{not} for acknowledging funding agencies.} 
  NYU Center for Data Science \\
  \texttt{guy.davidson@nyu.edu}
  \and
  Michael C.~Mozer \\
  Google Research\\
  University of Colorado at Boulder\\
  \texttt{mcmozer@google.com}}
  
% \author{Anonymous Authors}

\maketitle

\begin{abstract}

We explore the behavior of a standard convolutional neural net in a continual-learning setting that introduces visual classification tasks sequentially and requires the net to master new tasks while preserving mastery of previously learned tasks.  This setting corresponds to that which human learners face as they acquire domain expertise serially, for example, as an individual studies a textbook. Through simulations involving sequences of ten related visual tasks, we find reason for optimism that nets will scale well as they advance from having a single skill to becoming multi-skill domain experts. We observe two key phenomena. First, \emph{forward facilitation}---the accelerated learning of task $n+1$ having learned $n$ previous tasks---grows with $n$. Second, \emph{backward interference}---the forgetting of the $n$ previous tasks when learning task $n+1$---diminishes with $n$. 
Amplifying forward facilitation is the goal of research on metalearning, and attenuating backward interference is the goal of research on catastrophic forgetting. We find that both of these goals are attained simply through broader exposure to a domain. 

%{\color{blue} Third, we find that in this setting, our form of sequential training is far more efficient than simultaneous training, learning the sequence of tasks orders of magnitude faster.}
%We also observe intriguing behaviors that appear related to psychological phenomena, such as less forgetting with distributed practice, serial position effects, and  similarity-based interference.
\end{abstract}

% REMOVED FOR SPACE
%\section{Introduction}

In a standard supervised setting, neural networks are trained to
perform a single task, such as classification, defined in terms of a
discriminative distribution $p(y \,|\, x, \mathcal{D})$ for labels $y$ conditioned
on input $x$ and data set $\mathcal{D}$.  Although such models are useful
in engineering applications, they do not reflect the breadth required for general
intelligence, which includes the ability to select among many tasks.
%perform arbitrary tasks in a context-dependent manner.  
\emph{Multitask learning}  \citep{caruana1997}  is concerned with training models to
perform any one  of $n$ tasks, typically via a multi-headed neural network, where head $i$ represents the 
distribution $p(y_i \,|\, x,  \mathcal{D}_1, \ldots, \mathcal{D}_n )$.
Related tasks serve as regularizers on one another \citep{Caruana93,Ruder2017}.

%to extract shared structure \citep{Caruana93}, and as a regularization
%method to guide toward solutions helpful on a variety of problems \citep{Ruder2017}.

%Multitask learning is typically framed in terms of simultaneous training on all
%tasks, but humans and artificial agents operating in naturalistic settings more
%typically tackle tasks sequentially and need to maintain mastery of the
%previously learned tasks as they acquire the new one.  In this article, we
%perform an empirical investigation of training a single neural network 
%sequentially to acquire and maintain mastery of multiple tasks.

%Multitask learning is typically framed in terms of simultaneous training on all
%tasks, but 
%Humans and artificial agents operating in naturalistic settings more
%typically tackle tasks sequentially and need to maintain mastery of
%previously learned tasks as they acquire a new one. 

\emph{Continual} or \emph{lifelong learning} \citep{Thrun1996,Parisi2019} addresses a naturalistic variant 
in which tasks are tackled sequentially and mastery of previously learned tasks must be maintained while 
each new task is mastered.  Lifelong learning requires consideration of two issues:
\emph{catastrophic forgetting} \citep{McCloskey1989} and \emph{metalearning}
\citep{schmidhuber1987,bengioetal1991,Thrun1996},
Catastrophic forgetting is
characterized by a dramatic drop in task 1 performance
following training on task 2, i.e., the accuracy of the model $p(y_1 \,|\, x, \mathcal{D}_1\rightarrow\mathcal{D}_2)$
is significantly lower than accuracy of the model $p(y_1 \,|\, x,
\mathcal{D}_1)$, where the arrow
denotes training sequence.  
Metalearning aims to facilitate mastery on task $n$ from having
previously learned tasks $1, 2, \ldots, n-1$. Success in metalearning is
measured by a reduction in training-trials-to-criterion or an increase in
model accuracy given finite training for the $n$'th task, $p(y_n | x,
\mathcal{D}_1\rightarrow\ldots\rightarrow\mathcal{D}_n)$, relative to the
first task, $p(y_1 \,|\, x, \mathcal{D}_1)$.  

Researchers have proposed a variety of creative approaches---specialized mechanisms, 
learning procedures, and architectures---either for mitigating forgetting or for
enhancing transfer. We summarize these approaches in the next (``Related research'') section.
Although the literatures on catastrophic forgetting and metalearning have been 
considered separately for the most part, we note
that they have a complementary relationship.
Whereas catastrophic-forgetting reflects \emph{backward interference} of a new
task on previously learned tasks, metalearning reflects \emph{forward
facilitation} of previously learned tasks on a new task \citep{Lopez-Paz2017}.
Whereas catastrophic forgetting research has focused on the first task learned, metalearning 
research has focused on the last task learned. We thus view these two topics as endpoints
of a continuum. 

To unify the topics, we examine the continuum from the first task to
the $n$'th.  We train models on a sequence of
related visual tasks and investigate the consequences of introducing each new task
$i$.  We count the total number of training \emph{trials}---presentations of training examples---required to learn the $i$'th task
while maintaining performance on tasks $1\ldots i-1$ through continued practice.  Simultaneously, we
measure how performance drops on tasks $1\ldots i-1$ after introducing task $i$
and how many trials are required to retrain tasks $1\ldots i-1$.  We believe
that examining \emph{scaling} behavior---performance as a function of $i$---is
critical to assessing the efficacy of sequential multitask learning. Scaling
behavior has been mostly overlooked in recent deep-learning research, which is
odd considering its central role in computational complexity theory, and
therefore, in assessing whether existing algorithms offer any hope for
extending to human-scale intelligence.  

Surprisingly, we are aware of only one article \citep{Schwarz2018} that jointly considers forgetting
and metalearning through their scaling properties. However, this research,
like that in the catastrophic-forgetting and metalearning literatures,  suggests that specialized 
mechanisms are required for neural networks to operate in a lifelong learning setting.
The punch line of our article is that \emph{a standard neural network architecture
trained sequentially to acquire and  maintain mastery of  multiple visual tasks exhibits
faster acquisition of new knowledge and less disruption of previously acquired knowledge
as domain expertise expands.} We also argue that network learning and forgetting 
have an intriguing correspondence to the human and animal behavioral literature.

% GUY: we are not spacing practice in the usual sense. also spacing helps
% prevent forgetting over TIME.  we are not observing any temporal forgetting,
% only sequential interference. So let's leave out the spacing effect. I think
% you could get away with it, but not to a psychology audience.
%This analogy also helps explain our design
%choice of continuing to test on previous tasks, just as tests tend to contain
%materials from previous chapters, utilizing the effects of distributed practice
%(which relies on the spacing effect; cf. Bahrick et al., 1993; Russo et al.,
%1998).

\section{Related research}
A variety of mechanisms have been proposed to overcome catastrophic forgetting (for review, see \citep{Parisi2019}).
In addition to standard regularization techniques such as dropout \citep{Goodfellow2015}, specialized regularizers have been explored.
\citet{Kirkpatrick2017} introduce elastic weight consolidation, 
a regularizer which encourages stability of weights that most 
contribute to performance on previously trained tasks, and
\citet{Zenke2017} propose intelligent synapses that track their 
relevance to particular tasks.
Inspiration from biological systems has suggested models that
perform generative replay to recreate past experience in order to
retain performance on previously learned tasks \citep{Kamra2017},
and models that use consolidation mechanisms like those that take
place during mammilian sleep \citep{Kemker2018}.

%To overcome catastrophic forgetting, standard techniques such as dropout have been suggested \citep{Goodfellow2015},
%but most propose augmenting models with specialized mechanisms \citep[for review]{Parisi2019}.
%\citet{Kirkpatrick2017} introduce elastic weight consolidation, which adds a penalty to the model loss that
%encourages stability of weights  that most contribute to performance on previously trained tasks.
%\citet{Lopez-Paz2017} describe Gradient Episodic Memory, which retains examples of previous tasks 
%and minimizes the aforementioned negative backward transfer. \citet{Kemker2018} devise FearNet, a 
%neurally-inspired model with dual-memory design, using consolidation mechanisms modeled after mammalian 
%sleep consolidation. \citet{Zenke2017} are similarly biologically-inspired, motivating intelligent synapses 
%which track their relevance to particular tasks.
%%, not unlike the approach taken by \textcite{Kirkpatrick2017}. 
%\citet{Kamra2017} offer another dual-memory model, augmenting with a generative replay model able to 
%recreate past experiences and retain performance on previously learned tasks. 

To facilitate metalearning, mechanisms have been offered to encourage inter-task transfer,
such as MAML \citep{finn2017} and SNAIL \citep{mishra2018}. Other approaches employ recurrence
to modify the learning procedure itself \citep{Andrychowicz2016,Wang2017}.
\citet{Schwarz2018} construct a dual-component model consisting of a knowledge store of previously
learned tasks and an active component that is used to efficiently learn the current task. A consolidation
procedure then transfers knowledge from short- to long-term stores.

Within the catastrophic forgetting and metalearning literatures, some prior work focuses specifically on computer vision. \citet{Parisi2019} review existing vision-based benchmarks, including incremental training on MNIST \citep{LeCun1998MNIST}, the CUB-200 dataset \citep{Welinder2010}, and the CORe50 dataset \citep{Lomanco2017,Lomanco2019} for continual learning in an object recognition setting. Recently, \citet{lee2019overcoming} offered a novel method using distillation and confidence-based sampling to reduce catastrophic forgetting using unlabeled data. \citet{Aljundi2019} focus on eschewing the task identification and generate a self-supervised signal to recognize the transition between tasks. \citet{Stojanov2019} offer an incremental learning environment inspired by development, allowing for repeated encounters with the same objects from different perspectives.  

Despite the creativity (and success) of this assortment of methods, our concern centers on the fact that 
researchers assume the inadequacy of standard methods, and no attempt has been made to understand
properties of a standard architecture as it is trained sequentially on a series of 
tasks, and to characterize the extent of forgetting and transfer as more tasks are learned, \textit{while allowing continued practice of previously acquired tasks}.

% REMOVED FOR SPACE 
% The framework we propose will help us and others develop specialized mechanisms
% for attenuating backward interference and amplifying forward facilitation 
% But before we can focus on such specialized mechanisms, we focus on 
% plain old vanilla neural networks as a baseline. We find that they perform 
% surprisingly well.

\section{Methodology}

\begin{figure}[!tb]
\centering 
\includegraphics[height=.8in]{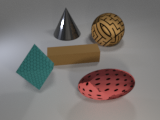}
\includegraphics[height=.8in]{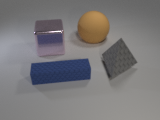}
\includegraphics[height=.8in]{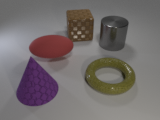}
\caption{Example training images}
\label{fig:examples}
\end{figure}

We train on visual tasks defined over images containing multiple colored, textured synthetic shapes (\autoref{fig:examples}). The tasks 
involve yes/no responses to questions about whether an image contains 
an object with certain visual properties, such as ``is there a red object?'' or
``is there a spherical object?'' We generate a series consisting of 10 \emph{episodes}; in each episode, a new task is introduced (more details to follow on the tasks). A model is trained from random initial weights
on episode 1, and then continues training for the remaining nine episodes.
% I removed the next sentences for space considerations
%We follow a model over the course of the ten episodes.
%Each simulation reported below follows a model trained once on all ten episodes of tasks in a single visual property.
In episode $i$, the model is trained on a mix of examples from tasks $1$ to $i$ until an 
accuracy criterion of 95\% is attained on a hold-out set \emph{for all tasks}. 
To balance training on the newest task (task $i$ in episode $i$) and retraining on previous tasks, we adapt the methodology of
%utilize a \emph{coreset}, inspired by 
\citet{Nguyen2018}: half the training set consists of examples from the newest task, and the other half consists of an equal number of examples from each of the previous tasks 1 through $i-1$. (In episode 1, only the single task is trained.) 
We evaluated multiple curricula, as characterized by the within-episode distribution of the number of training examples of each task. We found results to that we present below to be remarkably robust, and therefore we present
results in the main paper only for the balanced old/new split, but see Appendix A.2 for results from other curricula.

The same set of training images is repeated each epoch of training, but they are randomly reassigned to different tasks. 
Each epoch is divided into 30 training batches of 1500 examples.
Training continues for as many epochs as are required for all tasks in an episode to reach the accuracy criterion on the hold-out set. 
In each epoch, we roughly balance the number of positive and negative target responses for each task. 
% REMOVED FOR SPACE
%We turn now to details  of the images, tasks, and architecture.

%\begin{SCfigure} % [!tb]
%\end{SCfigure}

\emph{Image generation.} We leverage the CLEVR \citep{Johnson2017} 
%image generation
codebase to generate $160 \times 120$ pixel color images, each image containing 
4-5 objects that vary
along three visual \emph{dimensions}: shape, color, and texture. 
In each image, each object has a unique feature value on each dimension.
% MIKE REMOVED THE NEXT SENTENCE FOR SPACE
%: for example, no two objects in the same image will be blue.
%The original CLEVR dataset had only eight colors, three shapes, and two textures; 
We augmented CLEVR with additional features to ensure 10 values per 
dimension. (See supplementary material for details.)
We synthesized 45,000 images for a training set, roughly balancing the count of each feature across images, and
5,000 for a hold-out set used to evaluate the model after each epoch and determine when to transition to the next episode.
Each image could used for any task. 
Each epoch of training involved one pass through all images, with a random assignment of each image to a single task each epoch to satisfy the constraint on the distribution of tasks. 

\emph{Tasks.}
For each replication of our simulation, we select one of the three dimensions and
randomize the order of the ten within-dimension tasks.  To reduce sensitivity
of the results to the task order, we performed replications using a Latin square design \cite[ch. 9]{Bailey2008}, % \citep[][ch. 9]{Bailey2008}, 
guaranteeing that within a block of ten replications, each task will appear
in each ordinal position exactly once. We constructed six such Latin square blocks for each of the three dimensions (shape, color, and texture), resulting in 180 total simulation replications. Because we observed no meaningful qualitative differences across task dimensions (see supplementary material), the results we report below collapse across dimension.

\begin{figure}[bt]
% \begin{SCfigure*}[10][bt]
    \centering
\includegraphics[height=1.25in]{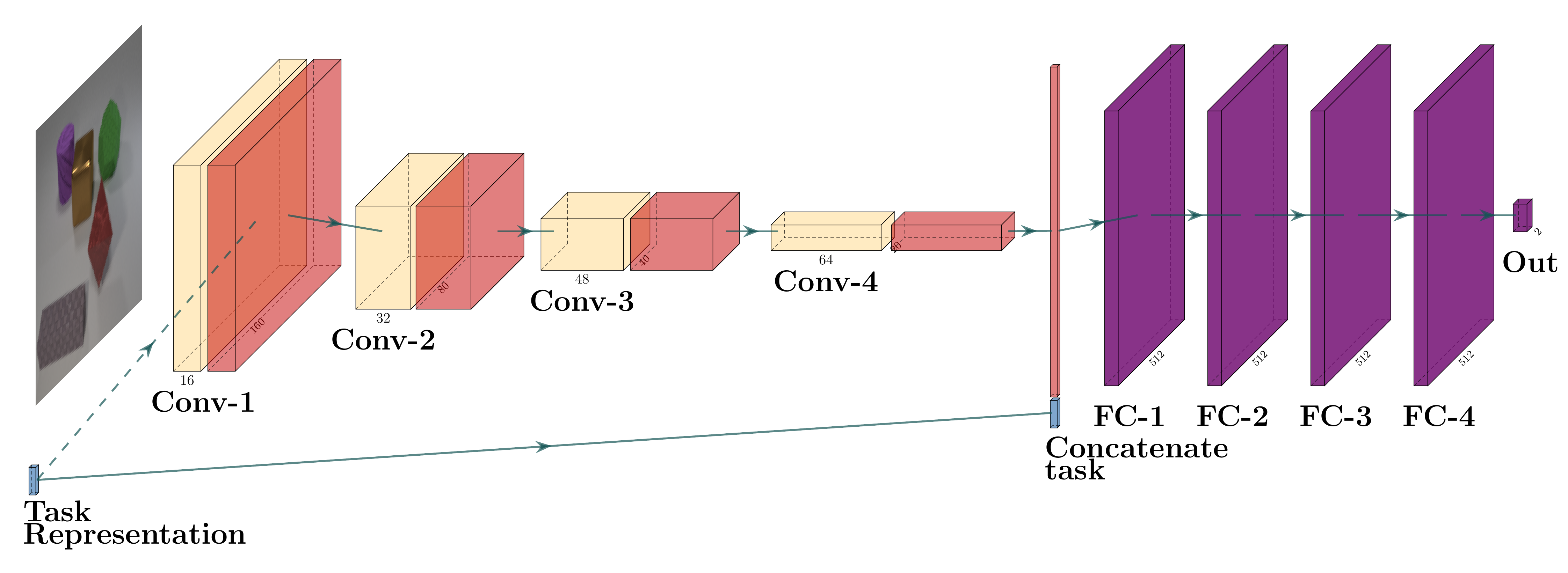}
\caption{Model architecture. Input consists of image and task representation. Dashed line from task representation to Conv-1 indicates optional task modulated visual processing, described under ``Task-modulated visual processing.''} % Section~\ref{sec:task-mod}.}
\label{fig:arch}
% \end{SCfigure*}
\end{figure}

\begin{figure*}[bt]
\centering
\includegraphics[width=\linewidth]{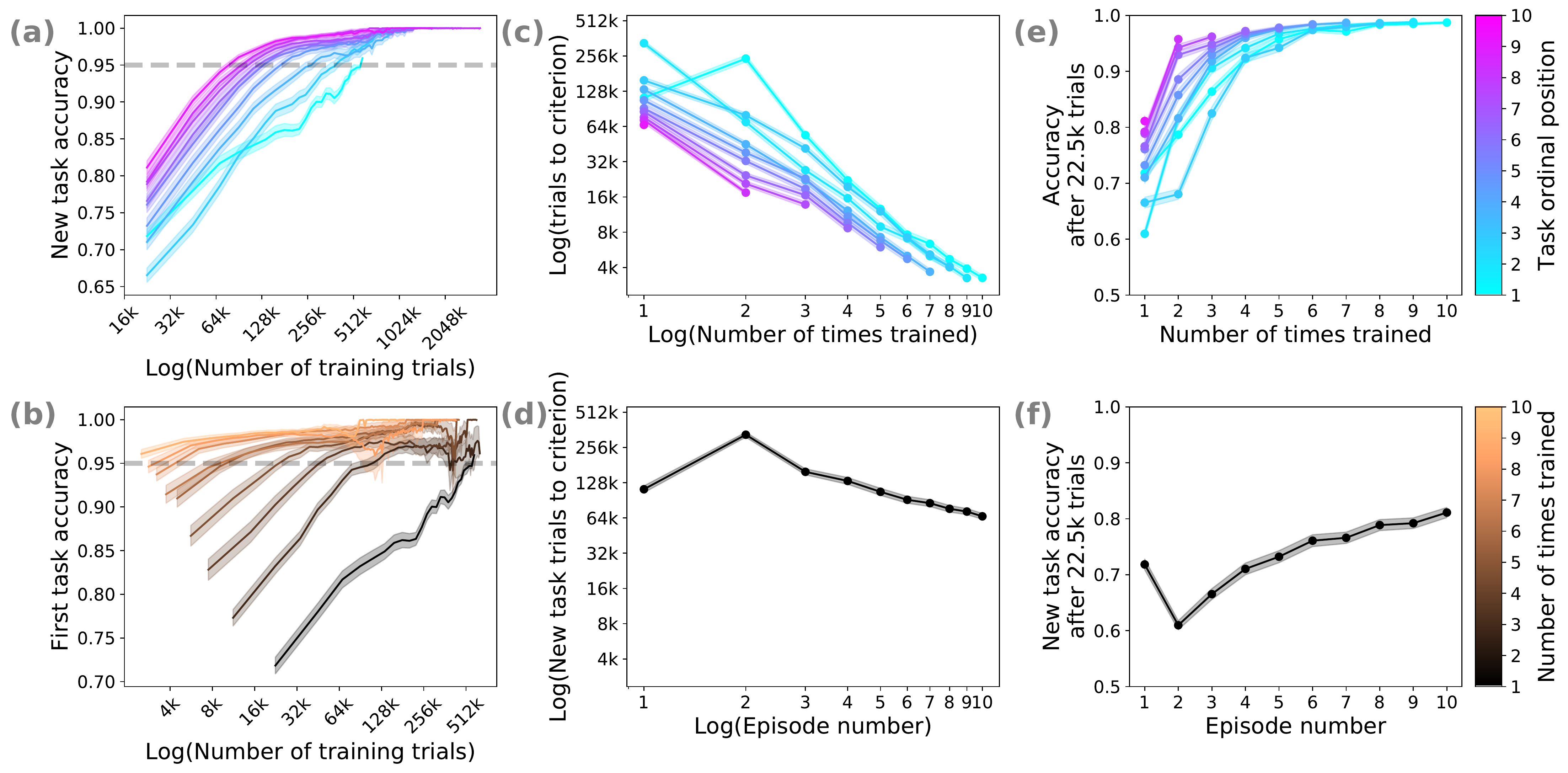}
\caption{\textbf{(a)} Hold-out set accuracy as a function of training trials (log scale) for a newly introduced task. Colored lines indicate task ordinal position (cyan = introduced in episode 1; magenta = introduced in episode 10). In all panels, the shaded region represents $\pm 1$ standard error of the mean. \textbf{(b)} Hold-out accuracy of the task introduced in episode 1 by number of times it is retrained (black = 1 time, copper = 10 times).
\textbf{(c)} Number of trials required to reach the accuracy criterion (log scale) as a function of
the number of times a given task is trained (also log scale). As in (a), the colors indicate task ordinal position (the episode in which a task is introduced).
\textbf{(d)} Similar to (c) but plotting only the new task introduced in a given episode.
\textbf{(e)} Hold-out accuracy attained after a fixed amount of training (22.5k trials) of a given task, graphed as a function of number of times a given task is trained. As in (a), the colors indicate the episode in which a task is introduced.
\textbf{(f)} Similar to (e) but plotting only the new task introduced in a given episode.
% GUY: i'm leaving this out because we've explained what we're showing; you're only reminding them that 22.5k epochs are reached after different numbers of epochs for the different points. Kind of subtle and not important to most readers.
%Due to how we allocate the coreset (see Methods), previously learned tasks receive less training per epoch as additional tasks are introduced, and therefore we plot the accuracy for an equal amount of training. 
}
\label{fig:results}
\end{figure*}

\emph{Architecture.}
Our experiments use a basic vision architecture with four convolutional layers followed by four fully connected layers (Figure~\ref{fig:arch}). The convolutional layers---with 16, 32, 48, and 64 filters successively---each have 3x3 kernels with stride 1 and padding 1,
followed by ReLU nonlinearities, batch normalization, and 2x2 max pooling. The fully-connected layers have 512 units in each, also with ReLU nonlinearities. All models were implemented in PyTorch \citep{Paszke2017} and 
trained with ADAM \citep{Kingma2015} using a learning rate of $0.0005$ and weight decay of $0.0001$. Note that our model is generic and is not specialized 
for metalearning or for preventing catastrophic forgetting.
Instead of having one output head for each task, we specify the task as a component of the input. Similar to Sort-of-CLEVR \citep{Santoro2017}, we code the task as a one-hot input vector. We concatenate the task representation to the output of the last convolutional layer before passing it to the first fully-connected layer.
We verified this architecture has the capacity to learn all thirty tasks (in all three dimensions) when trained simultaneously (see Appendix A.3). 
% REMOVED FOR SPACE: i don't think we need to say anything about NLP
%We presented the queries (which feature to answer about) as one-hot vectors,
%rather than in natural language, to isolate the visual processing from the
%natural language component that exists in most visual question answering (Antol
%et al., 2015) contexts

\section{Results}

\subsection{Metalearning}
\autoref{fig:results}a depicts hold-out accuracy for a newly introduced task as a function of the number of training trials. Curve colors indicate the task's ordinal position in the series of episodes, with cyan being the first and magenta being the tenth. Not surprisingly, task accuracy improves monotonically over training trials. But notably, metalearning is evidenced because the accuracy of task $i+1$ is strictly higher than the accuracy of task $i$ for $i>2$.
To analyze our simulations more systematically, we remind the reader that the simulation sequence 
presents fifty-five opportunities to assess learning: the task introduced in episode 1 (i.e., ordinal position 1) 
is trained ten times, the task introduced in episode 2 is trained nine times, and so forth, until the task introduced 
in episode 10, which is trained only once.  Figure~\ref{fig:results}c indicates, with one line per task,
the training required in a given episode to reach a hold-out accuracy of 95\%---the  dashed line in Figure~\ref{fig:results}a.
Training required per episode is plotted as a function of the number of times the task is retrained.
The downward shifting intercept of the curves for later tasks in the sequence indicates significantly easier learning and relearning.
Figure~\ref{fig:results}e shows an alternative view of difficulty-of-training by plotting accuracy after a fixed amount of (re)training. The conditions that require the least number of trials to criterion (Figure~\ref{fig:results}c) also achieve the highest accuracy after a small amount of training (Figure~\ref{fig:results}e).

\subsection{Catastrophic forgetting}
\autoref{fig:results}b shows the accuracy of the task introduced in the first episode ($y_1$) as it is retrained each episode.\footnote{The misalignment of the first point is due to the fact that the accuracy is assessed at the end of a training epoch, and each successive episode has fewer trials of task $y_1$ per epoch.}
The fact that performance in a new episode drops below criterion (the dashed line) indicates backward interference. However, there is a \emph{relearning savings}:
the amount of interference diminishes monotonically with the number of times trained. Notably, catastrophic forgetting of task 1 is essentially eliminated by
the last few episodes. Figure~\ref{fig:results}c shows very similar relearning savings for tasks 2-10 as for task 1. The roughly log-log linear
curves offer evidence of power-law decrease in the retraining effort required to reach criterion. 

Figure~\ref{fig:results} also reveals that the first two episodes are anomalous. Strong backward interference on task 1 is exhibited 
when task 2 is introduced (the crossover of the cyan curve in Figure~\ref{fig:results}c), a phenomenon that does not occur for
subsequent tasks. Similarly, strong forward interference on task 2 of task 1 is evident (slower learning for task 2 than for task 1 in 
Figure~\ref{fig:results}d), but tasks 3-10 are increasingly facilitated by previous learning.
These findings suggest that to understand properties of neural nets, we must look beyond training on 
just two tasks, which is often the focus of research in transfer learning and catastrophic forgetting.

\subsection{Resilience to forgetting}

\begin{figure}[bt]
% \begin{SCfigure*}[10][bt]
    \centering
\includegraphics[width=3.25in]{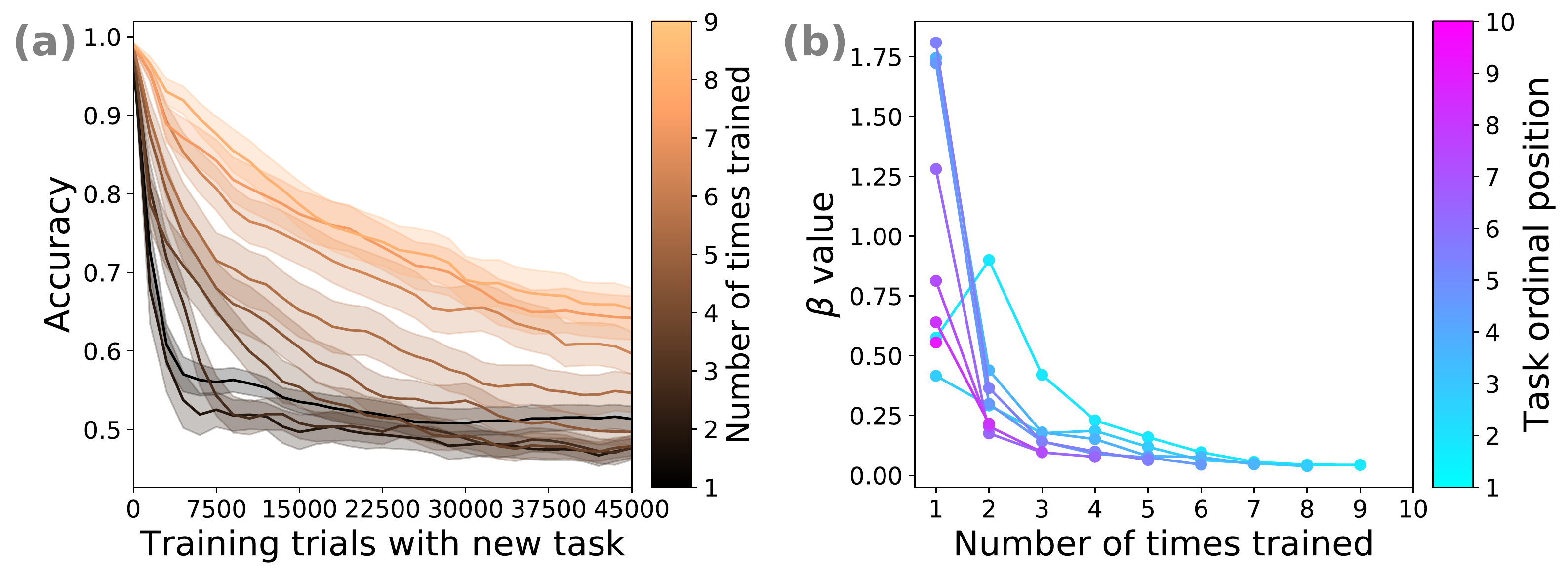}
\caption{Exploration of forgetting. 
% \textbf{(a)} Residual accuracy on task $i$ as the model is trained only on task $i+1$. 
%Colored lines indicate task ordinal position (cyan = introduced in episode 1; magenta = introduced in episode 9). 
%In all panels, the shaded region represents 
%$\pm 1$ standard error of the mean.  
\textbf{(a)} Residual accuracy of task 1 as the task introduced in episodes 2-10 is
trained (corresponding to 1-9 times that task 1 had previously been trained, with a black and copper for 1 and 9, respectively).
\textbf{(b)} The inferred exponential decay rate as a function 
of the number of times a task is trained.  
%As in (a),  color indicates task ordinal position (the episode in which a task  is introduced). 
% \textbf{(d)} Similar to (c) but graphed as a function of episode number. 
%As in (b), color indicates the number of times a task is retrained.
}
\label{fig:forgetting}
% \end{SCfigure*}
\end{figure}

%\begin{figure*}[bt]
%\centering
%\includegraphics[width=4.5in]{figures/forgetting_panel.pdf}
%\caption{Exploration of forgetting. \textbf{(a)} Residual accuracy on task $i$ as the model is
%trained only on task $i+1$. Colored lines indicate task ordinal position (cyan = introduced in 
%episode 1; magenta = introduced in episode 9). In all panels, the shaded region represents 
%$\pm 1$ standard error of the mean.  \textbf{(b)} Residual accuracy in episode $i$ of the task
%introduced in episode 1, as the model is trained only on task $i$.
%(black = task 1 has been trained 1 time previously, copper = 9 times previously).
%\textbf{(c)} The inferred forgetting decay rate as a function of the number of times a task
%is trained.  As in (a),  color indicates task ordinal position (the episode in which a task 
%is introduced). \textbf{(d)} Similar to (c) but graphed as a function of episode number. As in (b),
%color indicates the number of times a task is retrained.}
%\label{fig:forgetting}
%\end{figure*}
%
The fact that old tasks need to be retrained each episode suggests that
training on a new task induces forgetting of the old. However, because
we trained simultaneously on the old and new tasks, we have no
opportunity to examine forgetting explicitly. 
Inspired by results from the human learning literature \citep{Cepeda2008,Taylor2010}, we hypothesized that the forgetting rate would decrease with additional training. We devised a setting to examine this hypothesis by cloning weights at various points in the simulation
and examining
a different training trajectory moving forward. We took
the network weights at the
start of each episode $i$, at which point the network is at criterion
on tasks 1 through $i-1$. Then, instead of retraining on all $i$ tasks,
we train only on task $i$. We probe the network 
%after every batch  of 1500 training examples 
regularly to evaluate residual performance on old tasks.

%We performed an separate evaluation of how does forgetting of previously learned tasks scales as a function of the number of tasks and times trained. We performed this evaluation by training a model to successfully complete each episode as described above, but rather than continuing onto the next episode using the same sequential scheme, we trained the model \emph{only on a new task}. That is, we examined what would occur if we forewent the the coreset of training examples from previously learned tasks. We trained the new task for a single pass through the entire training set, and evaluated all previously learned tasks after every batch of 1500 training examples.

%{\color{red} \autoref{fig:forgetting}a depicts forgetting during episode $i$ of the
%task learned during the previous episode, $i-1$, when only task $i$ is
%trained. Regardless of a task's ordinal position in the series of
%episodes, indicated by curve color, forgetting is rapid and accuracy
%drops essentially to chance within 30k trials. There may be marginal
%sparing of the first (cyan) and next-to-last (magenta) tasks, but the
%effect is weak. }
%There appears to be little benefit for the most recently introduced task from previously learned tasks---while the earliest and latest tasks trained appear to learn marginally better, the effect is fairly negligible. 

\autoref{fig:forgetting}a depicts the time course of forgetting of the task introduced
in episode 1 on each subsequent episode. The black curve corresponds to episode 2 (task 1
has been trained only once previously) and the copper curve corresponds to episode 10
(task 1 has been trained 9 times previously). Task 1 becomes more robust to 
backward interference from the new task in later episodes, In episode $i$, task 1
has been (re)trained $i-1$ times previously, yielding a sort of spaced practice that
appears to cause the memory to be more robust. This result  is suggestive of the 
finding in human memory that interleaved, temporally distributed practice yields
more robust and durable memory \citep{Kang2014,Cepeda2008}.

Figure~\ref{fig:forgetting}a depicts only some of the 
forty-five opportunities we have to assess forgetting: we have one after the model learns a single 
task, two  after the model learns two, up to nine after the model learns the ninth task 
(for which we examine forgetting by training on the tenth and final task in the order).
To conduct a more systematic analysis, we fit the forgetting curves for each task $i$ in each 
episode $e > i$. The forgetting curve characterizes accuracy $a$ after  $t$ training batches of 1500 trials.
%corresponding to a single batch or 1500 training trials for the new task.
Accuracy must be adjusted for guessing: because our tasks have a baseline
correct-guessing rate of 0.5, we define 
$a = 0.5 + 0.5m$, to be the observed accuracy when memory strength 
$m$  lies between 0 (no task memory) and 1 (complete and accurate task memory).
We explore two characterizations of memory strength. The first is of
exponential decay, $m = \alpha \exp(- \beta t)$, where $\alpha$ is the
initial accuracy, $\beta$ is a decay rate, and $t$ is the number of intervening training batches.
The second is of power-law decay,
$m = \alpha (1 + \gamma t)^{-\beta}$, where $\gamma$ serves as a  timescale variable. 
This power-law decay curve is common in the psychological literature on forgetting 
\citep{Wixted2007} and has the virtue over $m=\alpha t^{-\beta}$ that it can characterize 
% REMOVED FOR SPACE: memory 
strength at $t=0$.

We fit the exponential and power-law functions separately to the data from each of the 45 model 
training points across 
67 replications of our experiment. Following \citet{Clauset2009}, we fit each form to the first 
half of the data, and assess it on the second half of the data. The power-law function obtains
a substantially lower MSE on the training data (power-law: 0.0045, exponential: 0.0198), the 
exponential function fit the held-out data better (power: 0.0232, exponential: 0.0192), and the 
exponential function offered a better fit on 24 of 45 training points of the model.
We therefore adopt the exponential-decay function and characterize decay by rate parameter $\beta$.

Figure~\ref{fig:forgetting}b presents the inferred decay rate $\beta$ for each of the 
forty-five model training points, presented in the style of Figures~\ref{fig:results}c,e. The
basic pattern is clear: additional practice yields a more durable memory trace, regardless of a
task's ordinal position. Further, with the exception of tasks 1 and 2, the forgetting rate of task 
$i$ on episode $i+1$ decreases with $i$, One is tempted to interpret this effect in terms of 
studies of human long-term memory, where \emph{serial position effects} are a robust 
phenomenon: Items learned early and late are preserved in memory better than items learned in 
between \citep{Glenberg1980}. Psychological studies tend to train to people only once per task \cite[e.g.,][]{Neumann1957} or multiple times on a single task \cite[e.g.,][]{Gerson1978}, so there are
no behavioral data concerning how serial position interacts with number of times trained, 
as we have in the simulation. There are a number of respects in which our simulation methodology 
does not align with experimental methodology in psychological studies, such as the fact that we
assess forgetting shortly after exposure, not at the end of a sequence of tasks. Nonetheless,
the correspondence between our simulation and human memory is intriguing. 

%We also see higher forgetting for the first task on the second time it is learned than the first time, offering further evidence to the catastrophic forgetting of the first task incurred by learning the second task. 

\begin{figure}[b!]
% \begin{SCfigure*}[10][b!]
\centering
\includegraphics[width=3.25in]{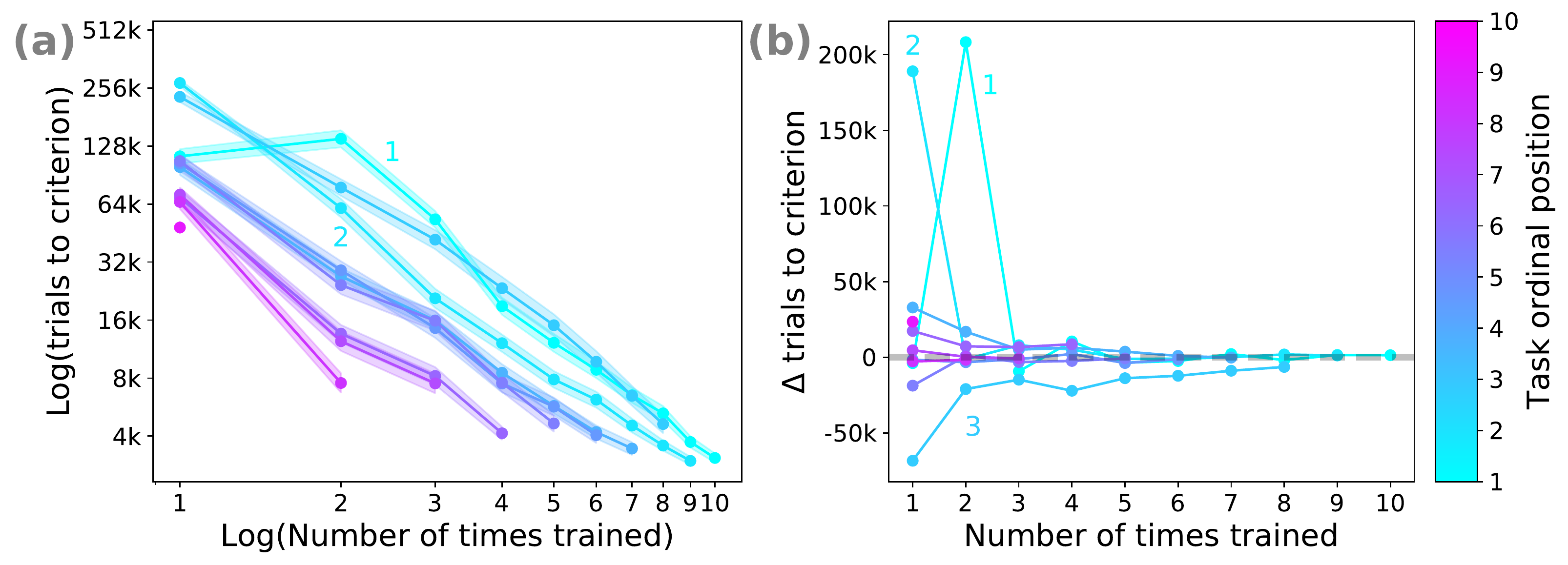}
\caption{Heterogeneous task sequences. 
\textbf{(a)} Number of trials required to reach the accuracy criterion versus number of times
a task is trained (cf. Figure~\ref{fig:results}c). The first two tasks are labeled by the numbers 1 and 2.
%(log scale) as a function of the number of times a given task is trained (log
%scale), corresponding to Figure~\ref{fig:results}c for homogeneous sequences. 
%  \textbf{(b)} Number of trials required to reach the accuracy criterion versus episode number (cf. Figure~\ref{fig:results}d).
% corresponding to Figure~\ref{fig:results}d for homogeneous sequences.
\textbf{(b)} Increase in number of trials required to reach accuracy criterion 
for homogeneous sequences compared to heterogeneous as a baseline.  Positive values indicate points 
learned faster in the heterogeneous condition, negative values in the baseline condition.
% \textbf{(d)} Similar to (c) but graphed as a function of episode number.
}
%with the line colors indicating---as in (b)---the number of times a task is retrained.}
%\caption{Heterogeneous task sequences. \textbf{(a)} Number of trials required to reach the
%accuracy criterion (log scale) as a function of the number of times a given task is trained (log
%scale), corresponding to Figure~\ref{fig:results}c for homogeneous sequences. 
%\textbf{(b)} Number of trials required to reach the accuracy criterion (log scale) as a 
%function of the episode number, corresponding to Figure~\ref{fig:results}d for
%homogeneous sequences.
%\textbf{(c)} The difference in the number of trials required to reach the accuracy criterion 
%for heterogeneous sequences versus homogeneous sequences.  Positive values indicate points 
%learned faster in the heterogeneous condition, negative values in the baseline condition.
%\textbf{(d)} Similar to (c) but graphed as a function of episode number with the line colors indicating---as in (b)---the number of times a task is retrained.}
\label{fig:heterogeneous}
% \end{SCfigure*}
\end{figure}

\subsection{Heterogeneous task sequences}
We noted two benefits of training on task sequences: reduced backward interference and increased
forward facilitation. We next try to better understand the source of these benefits. In particular,
we ask how the benefits relate to similarity among tasks.
Previously, we sampled tasks \emph{homogeneously}: 
all ten tasks in a sequence were drawn from a single dimension (color, shape, or texture). We
now explore the consequence of sampling tasks \emph{heterogeneously}: the ten tasks in a sequence
draw from all three dimensions. Each replication utilizes a single permutation of the three 
dimensions and samples the ten tasks cycling between the dimensions (four from the first, 
three from the other two). We employed a similar Latin square design \cite[ch. 9]{Bailey2008} to balance between 
the permutations, such that each block of six replications includes each permutation once.

% Do we want to also mention the log-log regression lines / power law curves here? For what it's worth, the regression line for the original data (Figure 2c) is ___ log(y) = -1.603 log(x) + 11.71 ___ and for Figure 4a it is ___ log(y) = -1.531 log(x) + 11.45 ___

Figure~\ref{fig:heterogeneous}a presents the results of 114 replications of the heterogeneous
sequences, nineteen  using each of the six dimension permutations. To better compare to the
homogeneous sequence results (Figure~\ref{fig:results}c), Figure~\ref{fig:heterogeneous}b
plots the \emph{increase} in number of trials to criterion with \emph{homogeneous} 
sequences compared to heterogeneous as 
a baseline. With several exception points, the differences are not notable,
suggesting that inter-tasks effects with heterogeneous sequences are similar to those with
homogeneous sequences. Thus, inter-task effects appear to be primarily due learning to process
visual images in general, rather than the specific task-relevant dimensions.
The two outlier points in Figure~\ref{fig:heterogeneous}b
concern the first two episodes: With heterogeneous training, the interference between tasks 1 and
2 nearly vanishes, perhaps because the resources and representations required to perform the two
tasks overlap less. One might have predicted just the opposite result, but apparently,
extracting information relevant for one visual dimension does not inhibit the construction
of  representations suitable for other dimensions. We argue that this finding
is sensible given that the dimensions of color, form and texture are fundamentally confounded in the input image: interpreting color and texture can require information about an 
object's shape, and vice-versa. 

%In fact, the result appears consistent with a finding from human memory---that reducing (semantic) similarity of items reduces interference among them \cite{Baddeley1966}.

% Conversely, with heterogeneous training, the forward facilitation between tasks 1 and 2 to task 3 is much weaker, to the extent that task 3 is learned faster in the homogeneous condition. This effect underscores the forward facilitation acquired in a domain by as few as two related tasks.

\subsection{Task-modulated visual processing} \label{sec:task-mod}
The architecture that we have experimented with thus far treats the convolutional layers as 
visual feature extractors, trained end-to-end on task sequences, but the convolutional layers 
have no explicit information about task; task input is provided only to the final layers
of the net. In contrast, processing in human visual cortex can be task
modulated \cite{Fias2002,Bracci2017}. % \cite[e.g.,]{Fias2002}. 
Perhaps modifying the architecture to provide task information
to convolutional layers would reduce inter-task interference. Along the lines of
\citet{Mozer2008}, we investigated a modified model using task-modulated visual processing, 
adopting a simpler approach than most existing architectures for conditional
normalization 
or gated processing \citep{Perez2017,Chen}. We consider task modulation via a task-specific learned bias 
for each channel in a convolutional layer. As before, task is coded as a one-hot vector.
We incorporate connections from the task representation to a convolutional 
layer (Figure~\ref{fig:arch}), with one \emph{bias} parameter for the Cartesian product of tasks and channels.
This bias parameter is added to the output of each filter in a channel before applying 
the layer nonlinearity.
%Our approach requires introducing a single additional layer, whose size depends only on the size of the encoded task representation and the number of filters in the layer we wish to modulate. 

\begin{figure}[bt]
\centering
\includegraphics[width=3.25in]{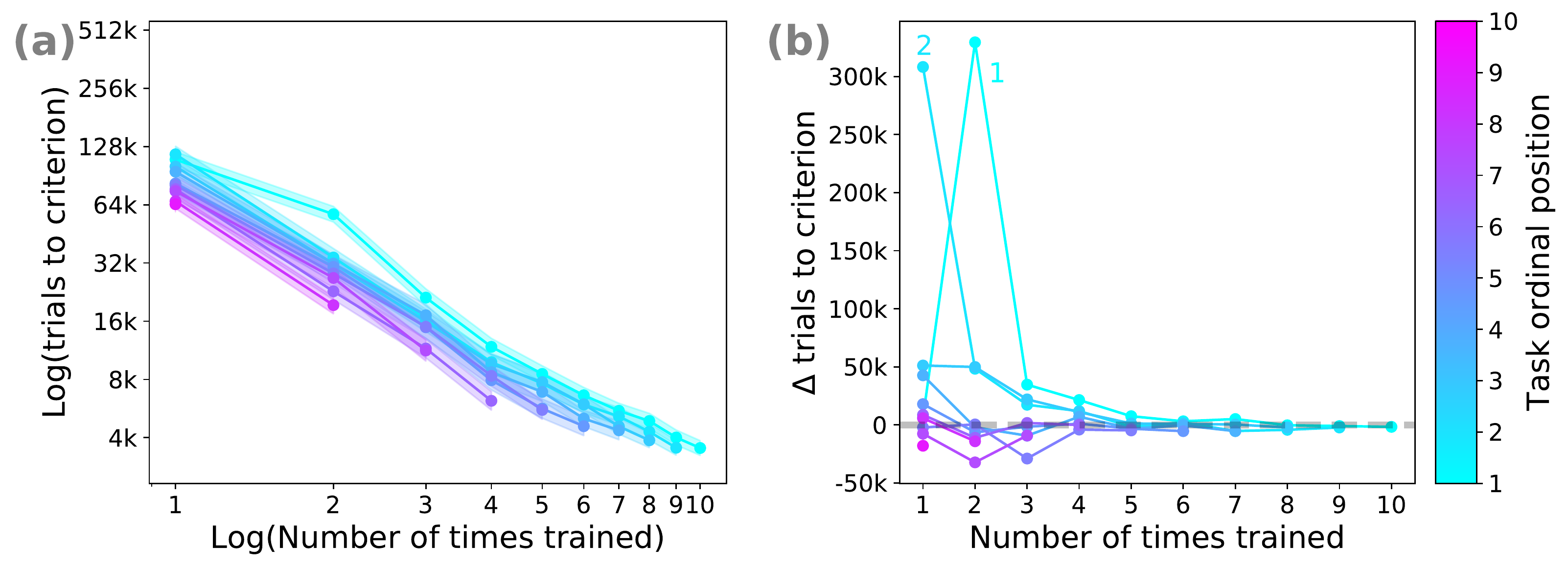}
\caption{Effect of modulating first convolutional layer with information about the current task.
\textbf{(a)} Number of trials required to reach the accuracy criterion versus number of times
a task is trained (cf. Figure~\ref{fig:results}c).
% \textbf{(b)} Number of trials required to reach the accuracy criterion versus episode
% number (cf. Figure~\ref{fig:results}d).
\textbf{(b)} Increase in number of trials required to reach accuracy criterion 
for non-task-modulated versus task modulated architectures. 
% \textbf{(d)} Similar to (c) but graphed as a function of episode number.
% \textbf{(e)} Effect of task modulation on trials-to-criterion for the first few episodes and
% tasks. The bulk of the effect is in episode 2 (graph is log scaled).
}

\label{fig:task-mod}
\end{figure}

We investigated task modulation at each of the four convolutional layers in our model. Because
the results of task modulation at the different layers are quite similar (see supplementary material),
we report the results of modulating processing at the first convolutional layer.
%, as it is furthest away from the original introduction of the task representation, and it was allowed to learn the fewest parameters (as the first convolutional layer had fewer filters than later layers). 
\autoref{fig:task-mod} depicts the results of three Latin square replications, yielding 
thirty simulations for each dimension, or ninety in total. Introducing task-based modulation 
allows the model to avoid catastrophic forgetting previously observed from learning 
the second task on the first, and to a lesser effect, improves performance in the 
third episode as well. As the model learns additional tasks, and continues retraining 
on the same tasks,  the benefits of task-modulation diminish rapidly 
(Figure~\ref{fig:task-mod}b), suggesting the primary benefit is in aiding early learning.  
We hypothesize that modulating visual processing with the task representation 
allows the model to learn flexible visual representations that produce less interference.

% \subsection{MAML}
% MAML \citep{finn2017} is designed to perform metalearning on a sequence of tasks in order to learn the next task in the sequence more efficiently.
% However, it is not designed to acquire mastery across multiple tasks simultaneously
% or to be cumulative in its knowledge.
% Nonetheless, there is a natural mapping from our domain to MAML. In our domain, we have a 10-step sequence of episodes, where tasks accumulate across episodes, resulting in $k$ tasks to be learned in episode $k$. MAML is also trained over a sequence
% of episodes, but we make a correspondence between one episode of MAML (the outer loop of the algorithm)
% to what we will refer to as a \emph{micro-episode} of our procedure.
% At the start of each micro-episode, we start with network weights $\bw$, and
% we draw a batch of 1500 examples which are 50\% from the newest task, 
% task $k$, and the remainder split evenly across the $k-1$ tasks. 
% (For task 1, all examples are from task 1.) From $\bw$, we 
% compute a gradient step based on the examples for each task $k$, and apply this 
% step separately to $\bw$, yielding $k$ copies of the weights, $\{\bw_1, ...,
% \bw_k\}$,  each specialized for its corresponding task. We then draw a second batch 
% of 1500 examples and perform a metatraining step, as described in MAML. This 
% involves computing the gradient with respect to $\bw$ for the new examples of
% task $k$ based on the weights $\bw_k$. Following MAML, we then update $\bw$, and 
% proceed to the next micro-episode until our training criterion is attained.
\subsection{Comparison to MAML}

The results we have presented thus far serve as a baseline against which one can compare any method specialized to reduce forgetting
or boost transfer. We conducted comparisons to several such methods, and in this section we report on experiments with 
\emph{model-agnostic metalearning} or \emph{MAML} \citep{finn2017}.
MAML is designed to perform metalearning on a sequence of tasks in order to learn the next task in the sequence more efficiently.
However, it is not designed for our continual-learning paradigm, which requires preservation of mastery for previous tasks.
We explored two variants of MAML adapted to our paradigm. We report here on the more successful of the two (see supplementary material for details).

Our paradigm is based on a series of 10 episodes where tasks accumulate across episodes. MAML is also trained
over a series of episodes, but we make a correspondence between one episode of MAML---the outer
loop of the algorithm---and what we will refer to as a \emph{micro-episode} of our paradigm, which corresponds to
a single batch in our original training procedure.
Each micro-episode starts with network weights $\bw$, and
we draw a half-batch of 750 examples (compared to 1500 in the original setting) of which 50\% are from the newest task, 
and the remainder are split evenly across the previous tasks.
(For task 1, all examples are from task 1.) From $\bw$, we 
compute a gradient step based on the examples for each task, and apply this 
step separately to $\bw$, yielding $i$ copies of the weights in episode $i$, $\{\bw_1, ...,
\bw_i\}$,  each specialized for its corresponding task. We then draw a second half-batch
of 750 examples and perform a metatraining step, as described in MAML.
Metatraining involves computing the gradient with respect to $\bw$ for the new examples of
each task $k$ based on the weights $\bw_k$. Following MAML, we then update $\bw$, and 
proceed to the next micro-episode until our training criterion is attained. 
Having halved the batch size, we doubled the learning rate from $0.0005$ in the original setting to $0.001$ for both of MAML's learning rates. 
%The model architecture used with MAML is otherwise identical to the one used without it. 
Model details are otherwise identical to the base model.

\begin{figure}[tb!]
\centering
\includegraphics[width=3.25in]{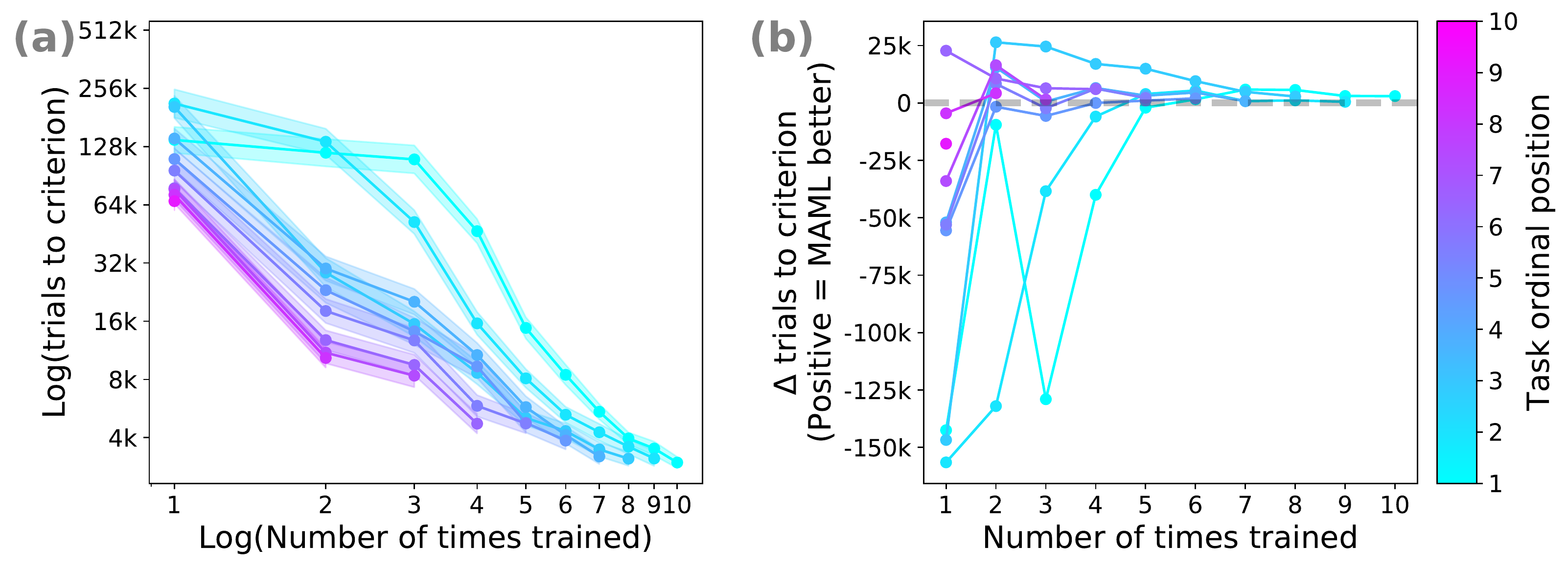}
\caption{MAML. 
\textbf{(a)} Number of trials (inner and outer loops combined) required to reach the accuracy criterion versus number of times
a task is trained (cf. Figure~\ref{fig:results}c). 
\textbf{(b)} Increase in number of trials required to reach accuracy criterion 
for training without MAML compared to utilizing MAML.  Negative values indicate slower learning
with MAML than the base model.}
\label{fig:maml}
\end{figure}

Over 90 replications (30 per dimensions), we find that the performance of our MAML variant is qualitatively similar to that of our base model (compare
\autoref{fig:maml}a and \autoref{fig:results}c).
% TO GUY FROM MIKE: i'm not seeing a need to say the following.
%, providing additional confidence that our result are robust to variations in the training procedure
However, quantitatively, the MAML-based method requires more trials to reach criterion on expectation: Figure~\ref{fig:maml}b shows the relative number of trials to criterion, where negative indicates that MAML is \emph{worse} than our base model.
Apparently the cost of splitting the data and devoting half to meta-training does not outweigh the benefit of meta-training.
%This second finding suggests that MAML might not be optimal for settings which require both maintaining previous tasks and efficiently learning new ones, as we did not see consistent evidence for increased forward facilitation using MAML, but rather generally worse results.

\section{Discussion}

We explored the behavior of a standard convolutional neural net for classification in a continual-learning setting that introduces visual tasks 
sequentially and requires the net to master new tasks while preserving mastery of previously learned tasks.  
This setting corresponds to that which human learners naturally face as they become domain experts. 
For example, consider students reading a calculus text chapter by chapter. 
Early on, engaging with a chapter and its associated exercises results in forgetting of previously mastered material.  
However, as more knowledge is acquired, students begin to scaffold and link knowledge and eventually are able to integrate the new material with the old.
As the final chapters are studied, students have built a strong
conceptual framework which facilitates the integration of new material with little disruption of the old.  
These hypothetical students behave much like the net we studied in this article.

We summarize our novel findings, and where appropriate, we link these findings more concretely to the literature on human learning.

\begin{enumerate}
    \item Metalearning (forward facilitation) is observed once the net has acquired sufficient domain expertise. In our paradigm, 
    `sufficient expertise'  means having mastered two tasks previously.
    Metalearning is demonstrated when training \emph{efficiency}---the number of trials needed to reach criterion---improves with each
    successive task
    (Figures~\ref{fig:results}d,f). Metalearning occurs naturally in the model and does not require specialized mechanisms. Indeed, incorporating a 
    specialized mechanism, MAML, fails to enhance metalearning in our continual-learning paradigm.
    
    \item Catastrophic forgetting (backward interference) is reduced as the net acquires increasing domain expertise (i.e., as more related tasks are learned).
    In Figure~\ref{fig:results}c, compare tasks introduced early (cyan) and late (magenta) in the sequence, matched for number of
    times they have been trained (position on the abscissa). Retraining efficiency improves for tasks introduced later in the task sequence, indicating
    a mitigation of forgetting. 
    Note that the number of trials to relearn a skill is less than the number of trials required to initially learn a skill
    (the exception being task 1 in episode 2). This \emph{relearning savings} effect has long been identified as a characteristic of human memory
    \citep{Ebbinghaus}, as, of course, has the ubiquity of forgetting, whether due to the passage of time \cite{Lindsey2014} or to backward interference
    from new knowledge \citep{Osgood1948}.
    % REMOVED FOR SPACE ,Postman1961}. 
    
    \item The potential for catastrophic forgetting (backward interference) is also reduced each time a task is relearned, as indicated by the monotonically decreasing
    curves in Figure~\ref{fig:results}c and by the change in forgetting rates in Figure~\ref{fig:forgetting}. A task that is practiced over
    multiple episodes receives \emph{distributed} practice that is \emph{interleaved} with other tasks. The durability of memory with
    distributed,
    interleaved practice is one of the most well
    studied phenomena in cognitive psychology
    \citep{Kang2014,Cepeda2008,Taylor2010,Birnbaum2013} and perceptual learning \citep{Szpiro2014},
    % REMOVED FOR SPACE,Szpiro2014}.
    helping human learners distinguish between categories, presumably by helping distinguish within- and between-category differences. 
    
    \item Training efficiency improves according to a power function of the number of tasks learned, controlling for experience on a task (indicated
    by the linear curve in Figure~\ref{fig:results}d, plotted in log-log coordinates), and also according to a power function of the amount of training a given task has received, controlling for number of tasks learned (indicated by the linear curves in Figure~\ref{fig:results}c). Power-law learning is a robust characteristic of human skill acquisition, observed on a range of behavioral measures \citep{Newell1980,Donner2015}.
    
    \item Forward facilitation and reduction in backward interference 
    is observed only after two or more tasks have been learned. This pattern can be seen by the nonmonotonicities in the curves of Figures~\ref{fig:results}d,f and in the crossover of curves in Figures~\ref{fig:results}c,e.
    Catastrophic forgetting is evidenced primarily for task 1 when task 2 is learned---the canonical case studied in the literature. However, the net becomes more robust as it acquires domain expertise, and eventually the relearning effort becomes negligible (e.g., copper curves in Figure~\ref{fig:results}b). The anomalous behavior of task 2 is noteworthy, yielding a transition behavior perhaps analogous 
    to the ``zero-one-infinity'' principle \cite{MacLennan1999}.
    
    \item Catastrophic forgetting in the second episode can be mitigated in two different ways:
first, by choosing tasks that rely on different dimensions (Figure~\ref{fig:heterogeneous}); and 
second, by introducing task-based modulation of visual processing (Figure~\ref{fig:task-mod}).
We conjecture that both of these manipulations can be characterized in terms of reducing the
similarity of the tasks. In human learning, reducing (semantic) similarity
reduces interference \cite{Baddeley1966}, and top-down, task-based signals interplay with perceptual learning \citep{Seitz2005,Byers2012}, as early in processing as primary visual cortex (V1) \citep{Li2004,Schafer2007}. Our results imply that inter-task similarity creates higher levels interference for the first few tasks acquired, compared to training on a diverse set of tasks, without conferring additional benefits on facilitation of tasks introduced later. Consequently, training a model on diverse tasks might reduce forgetting without reducing forward facilitation. Task-based modulation serves as another promising avenue to reducing interference by similar tasks, which we hope to continue exploring in future work.

% MIKE TO GUY: I REMOVED THIS BULLET PER OUR DISCUSSION
    %{ \color{blue} \item  We discover than in this setting, which allows explicit practice of previously learned tasks, sequential learning is orders of magnitude faster than simultaneous learning. 
    %This occurs even with a random curriculum (task introduction ordering), without attempting to optimize for difficulty. }
    
    %Forgetting appears to primarily be a function of times retrained on the same task, rather than of acquiring domain knowledge by learning additional tasks. Additionally, the decay rates for tasks learned once follow a non-monotonic pattern, with forgetting peaking not after one or two tasks, but rather after a few more, before starting to slowly improve with additional domain experience. 
    
\end{enumerate}
We are able to identify these intriguing phenomena because our simulations examined \emph{scaling behavior} and not just effects of one task on a second---the typical case for studying catastrophic forgetting---or the effects of many tasks on a subsequent task---the typical case for metalearning and few-shot learning. Studying the continuum from the first task to the $n$'th is quite revealing. 

We find that learning efficiency improves as more tasks 
are learned. Although MAML produces no benefit 
over the standard architecture that served as our baseline,
we have yet to explore other methods that are explicitly designed to 
facilitate transfer and suppress interference
\citep{mishra2018,Kirkpatrick2017,Lopez-Paz2017}.
The results presented in this article serve as a baseline to assess the benefits of specialized methods. A holy grail of sorts would be to identify methods that achieve \emph{backward facilitation}, where training on later tasks improves performance on earlier tasks, and \emph{compositional generalization} \citep{Fodor1988,Fodor2002,Lake2018,Loula2018}, where learning the interrelationship among earlier tasks allows new tasks to be performed on the first trial. Humans demonstrate the former under rare conditions \citep{Ausubel1957,Jacoby2015}; the latter is common in human behavior, as when individuals are able to perform a task immediately from instruction.

%The psychology literature discusses these effects as retroactive and proactive interference. Catastrophic forgetting, which we suggest evaluating as backward interference, is an example of retroactive interference, whereas metalearning, which we examine as forward facilitation, is an avoiding proactive interference, or proactive facilitation. An alternative unifying terminology, offered by \citet{Lopez-Paz2017}, suggests studying these as backward and forward transfer, although we prefer the distinction between the effects we tend to observe: the effect of multitask training tends to be negative looking backward, and positive looking forward. Term them as you may, we believe there is value in studying these properties jointly, on the continuum of multitask learning, rather than as two separate fields of inquiry. A holy grail of sorts would be to devise models that demonstrate backward facilitation, that training on later tasks improving performance on previously learned tasks, as human learners demonstrate in some conditions \citep{Ausubel1957,Jacoby2015}.

An exciting direction for future research concerns optimizing curricula for
continual learning. Our initial approach was
inspired by best practices of the science of learning
literature \citep{Weinstein2018}. Our hope is that
investigations of networks may in turn provide
helpful guidance for improving curricula for human learners.  Toward this goal, it is encouraging
that we observed more than superficial similarities
between human and network continual learning.

%{ \color{blue} 
%Instead, we offer a systematic investigation of learning in a more human-like fashion: rehearsing previously learned tasks while acquiring new knowledge.
%The science of learning literature (see review by \citep{Weinstein2018}) offers robust evidence for several best practices for human learners, which help inspire the design of our sequential multitask learning setting.
%Periodic review of previously learned knowledge \citep{Lindsey2014} is known to help students retain and master knowledge; we utilize a coreset (modified from \cite{Nguyen2018}) of examples from previously learned tasks in order to provide such practice. 
%
%
%However, rather than opt for the standard supervised learning paradigm of introducing (and interleaving) all tasks simultaneously, we build this interleaving gradually, allowing the model to acquire one task before progressing to the next one. 
%We build on these ideas in designing our curricular method, but we leave it to future work to investigate more optimal instantiations of these ideas. }
%
\clearpage
% {\tiny
% \bibliographystyle{ieee_fullname}
\bibliographystyle{IEEEtranN}
\bibliography{scaling} % .bib file

\clearpage
\appendix

\def \suppfigwdith{0.8} %{1.0}

%\begin{document}
%
%\title{Supplementary Information}

\section{Supplementary materials}
\subsection{Feature values}

The ten colors used in our experiments are:
gray, red, blue, green, brown, purple,
magenta, yellow, orange, pink. 
The ten shapes are: cube, sphere, cylinder,
pyramid, cone, torus, rectangular box, ellipsoid, octahedron, dodecahedron.
And the ten textures are:
metal, rubber, chainmail, marble, maze, metal weave, polka
dots, rug, bathroom tiles, wooden planks.
See additional example images below:

\begin{figure}[h]
\centering 
\includegraphics[height=1in]{figures/examples/CLEVR_new_000005.png}
\hspace{.05in}
\includegraphics[height=1in]{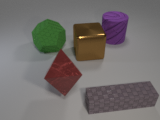}
\hspace{.05in}
\includegraphics[height=1in]{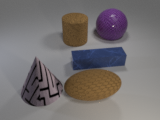}
\hspace{.05in}
\includegraphics[height=1in]{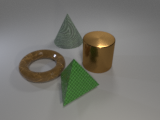}
\hspace{.05in}
\includegraphics[height=1in]{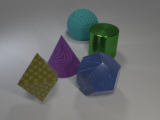}
\hspace{.05in}
\includegraphics[height=1in]{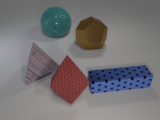}
\hspace{.05in}
\includegraphics[height=1in]{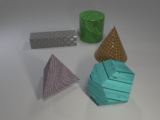}
\hspace{.05in}
\includegraphics[height=1in]{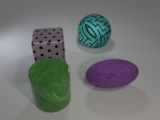}
\caption{Additional example training images}
\label{fig:additional-examples}
\end{figure}

\clearpage

\subsection{Curriculum ablations}

To explore the sensitivity of our results to our formulation of the coreset as a curriculum, we devised two alternative approaches, beyond the baseline approach reported of allocating half of the examples in each epoch to the newest task, and splitting the remaining examples evenly between previously learned tasks.
With all curricula explored below, we make an effort to attain the same balance both within each epoch as a whole, and within each batch in the epoch. 
We also maintained the same evaluation protocol as with the baseline curriculum to make sure the results are comparable. 

\subsubsection{Ratio curriculum} 
In this curriculum, we again allocate half of the examples in each epoch to the newest task. 
We split the remaining half such that the ratio between examples allocated to task $k$ and task $k + 1$ is $1:\kappa$, where $\kappa \geq 1$ is a free parameter.
This allows newer tasks, which the model learned fewer times, to receive proportionally more instruction than previous tasks. 
Note that for the first two episodes, this behaves identically to the baseline curriculum we investigated.
We explored two settings of $\kappa$: the first, $\kappa = 1.24$, we approximated from the data collected under our baseline curriculum, and a second, $\kappa = 1.5$, we chose to explore the sensitivity to this parameter. 

For example, with $\kappa = 1.25$, the third episode, the first task receives 10000 examples, the second 12500, and the third, and newest task, receives its 22500 examples in each epoch.
For another example, in the sixth episode, the tasks are allocated the following numbers of examples per epoch: 2741, 3427, 4284, 5355, 6693, and 22500.
With $\kappa = 1.5$, the third episode allocations are 9000, 13500, and 22500, and the sixth episode training example allocations are 1706, 2559, 3839, 5758, 8638, and 22500.

\subsubsection{Power curriculum} 
In this curriculum, unlike the baseline and the ratio curriculum we \textit{omit} the allocation of half of the examples to the newest task.
In episode $k$, for task $1, 2, \ldots t, \ldots, k$, we first compute unnormalized proportions using a power function: task 1 receives $\rho_1 = k^{-\alpha}$, task 2 receives $\rho_2 = {k - 1}^{-\alpha}$, and so on, such that each task $t$ receives a proportion of $\rho_t = (k - t + 1)^{-\alpha}$ of the examples. 
We then normalize the proportions, such that each task receives $p_t = \frac{\rho_t}{\sum_{i=1}^k \rho_i}$ of the 45000 total training examples per epoch. 
As with the ratio curriculum, we first estimated $\alpha$ from the baseline curriculum data, arriving at $\alpha = 1.14$, and then explored an alternative setting of $\alpha = 2$. 
For example, with $\alpha = 1.14$, in the third episode, task 1 receives 7394 training examples, task 2 receives 11738 examples, and task 3 (the newest task) receives 25868 examples. 
In the sixth episode, tasks 1-6 are allocated the following numbers of examples: 2611, 3215, 4146, 5755, 9137, and 20136.
With $\alpha = 2$, in the third episode we allocate 3674, 8265, and 33061 training examples to tasks 1-3 respectively, and in the six episode, we allocate tasks 1-6 838, 1207, 1886, 3353, 7543, and 30173 training examples respectively. 

\subsubsection{Results}
\begin{figure*}[b!]
\centering
\includegraphics[width=\suppfigwdith\linewidth]{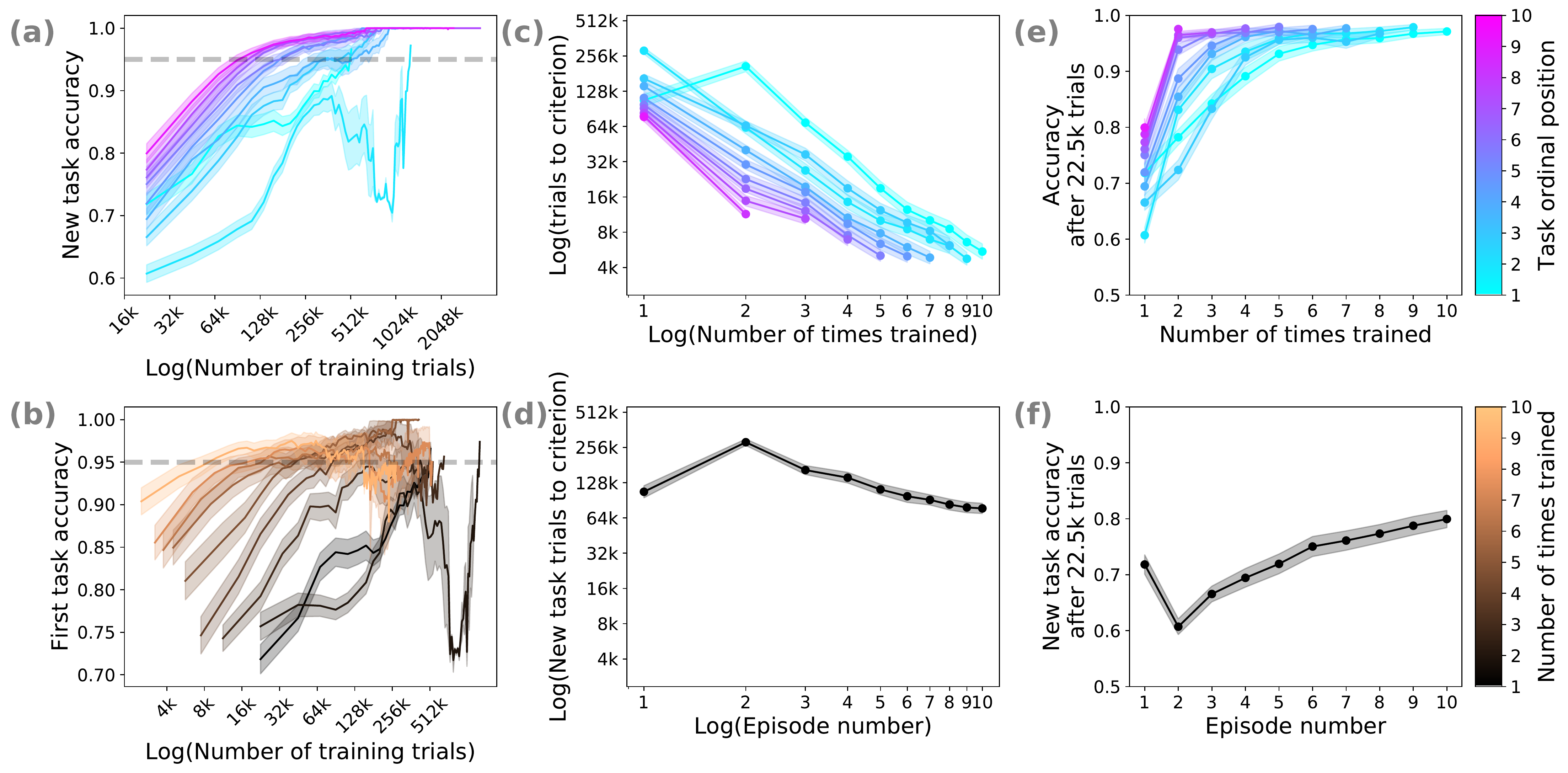}
\caption{Reproduction of \autoref{fig:results} with the data from the ratio curriculum using a ratio of $\kappa = 1.25$ approximated from the baseline curriculum data.}
\label{fig:appendix-curricula-ratio}
\end{figure*}

\begin{figure*}[b!]
\centering
\includegraphics[width=\suppfigwdith\linewidth]{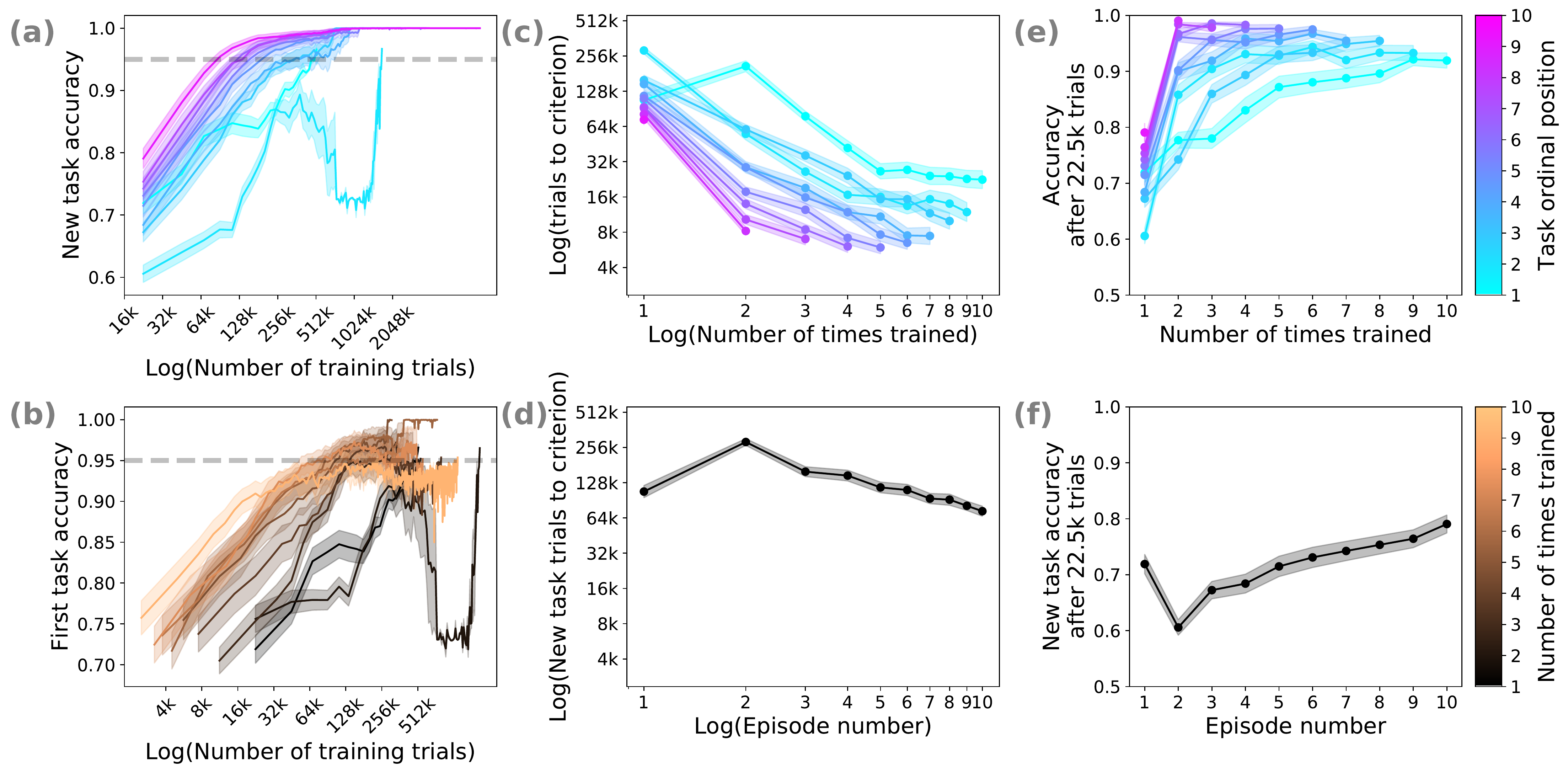}
\caption{Reproduction of \autoref{fig:results} with the data from the ratio curriculum using a ratio of $\kappa = 1.5$ designed to explore the sensitivity to the choice of parameter.}
\label{fig:appendix-curricula-ratio-1-5}
\end{figure*}

\begin{figure*}[b!]
\centering
\includegraphics[width=\suppfigwdith\linewidth]{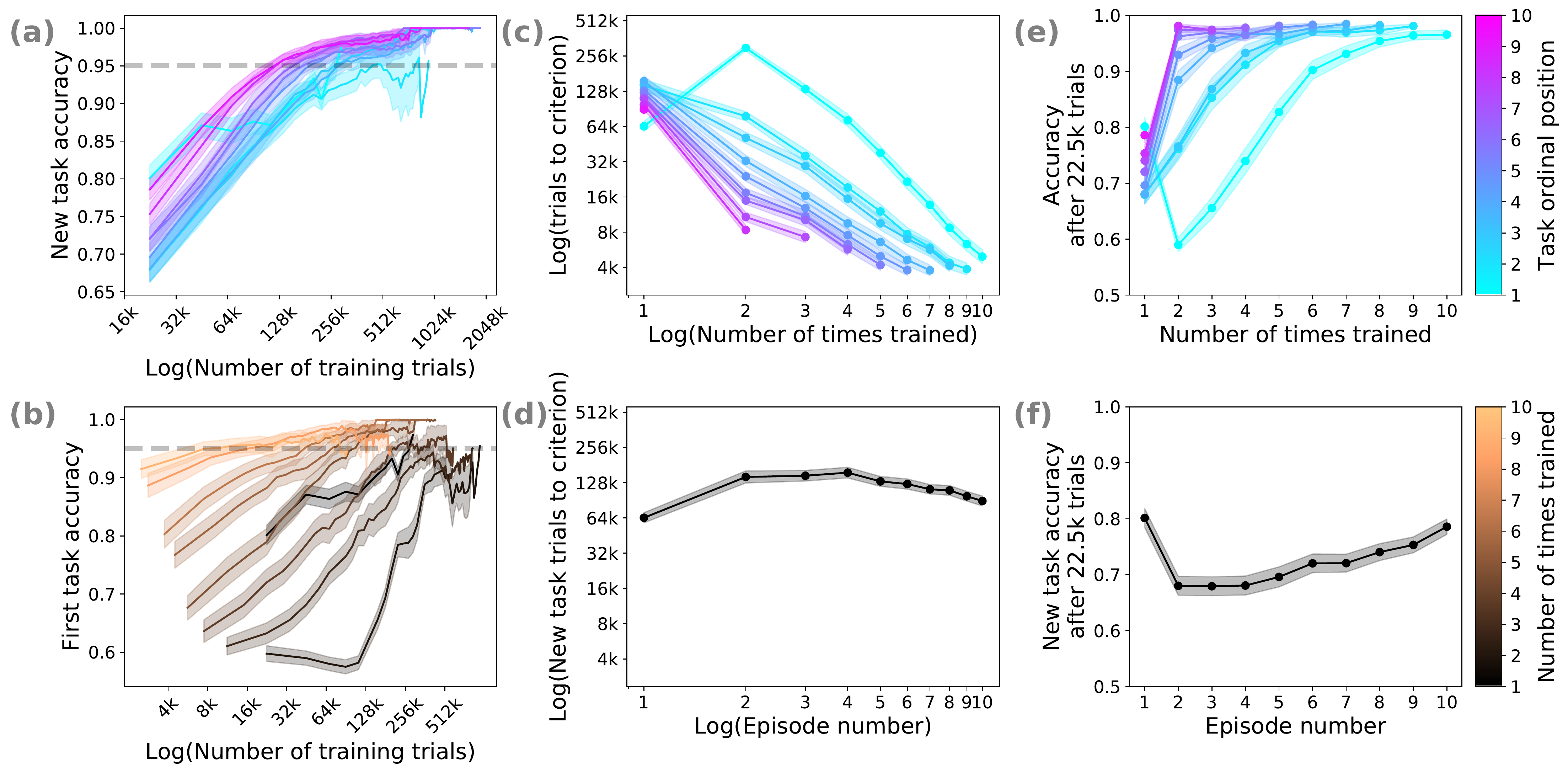}
\caption{Reproduction of \autoref{fig:results} with the data from the power curriculum using an exponent of $\alpha = 1.14$ approximated from the baseline curriculum data.}
\label{fig:appendix-curricula-power}
\end{figure*}

\begin{figure*}[b!]
\centering
\includegraphics[width=\suppfigwdith\linewidth]{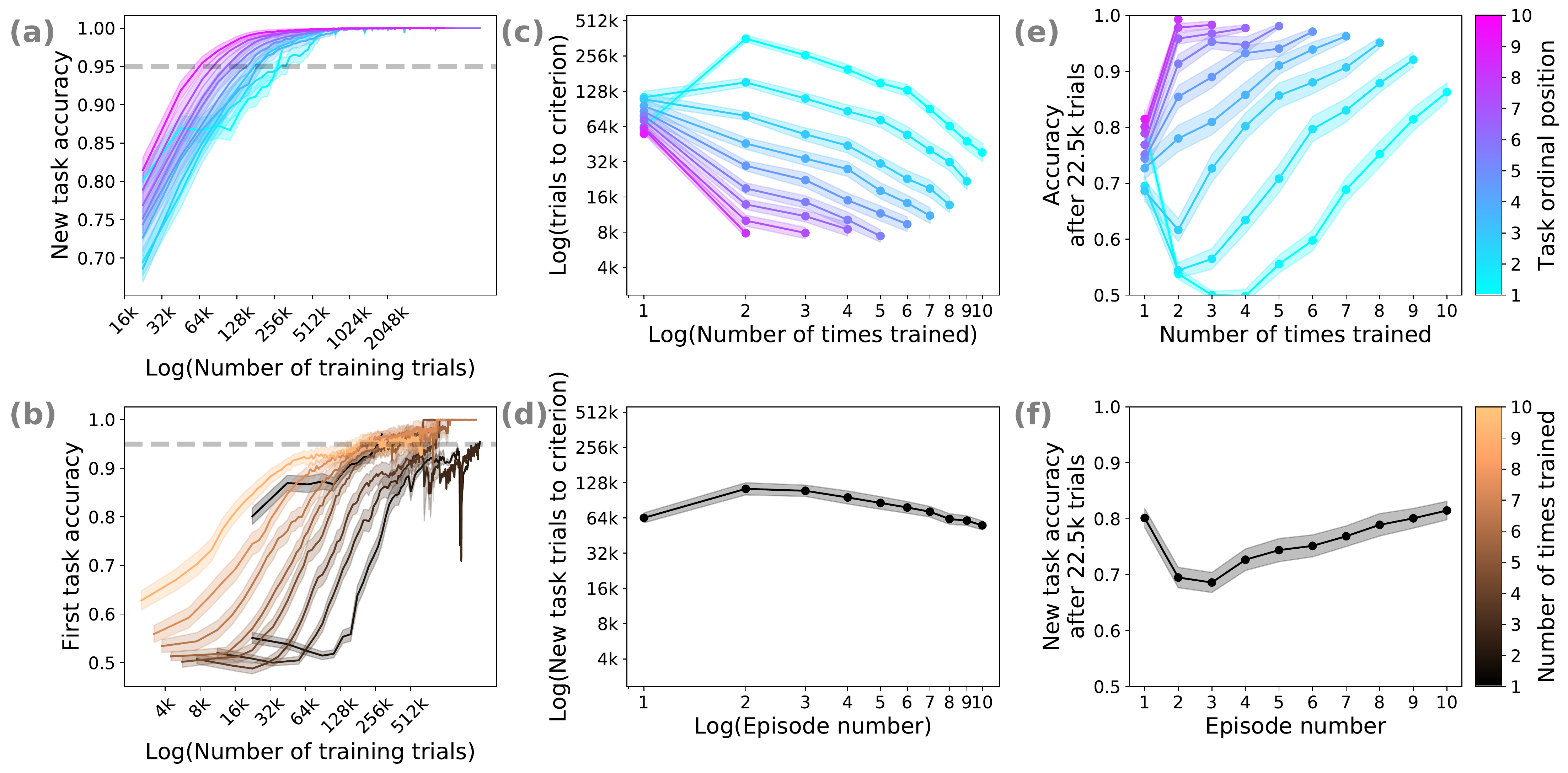}
\caption{Reproduction of \autoref{fig:results} with the data from the power curriculum using an exponent of $\alpha = 2$ designed to explore the sensitivity to the choice of parameter.}
\label{fig:appendix-curricula-power-2}
\end{figure*}

We reproduced \autoref{fig:results} with the data from running two Latin square replications in each dimension (resulting in sixty total runs) utilizing each curriculum.
The results in panels (a) and (b) are noisier, especially for above 256k examples, as most runs finish sooner, and thus there is less data in in that part of the plot.
Qualitative, we find panels (c)-(f) quite similar to the baseline curriculum, especially for the two data-optimized versions of the curriculum (\autoref{fig:appendix-curricula-ratio} and \autoref{fig:appendix-curricula-power}), and slightly less so for the misspecified versions (\autoref{fig:appendix-curricula-ratio-1-5} and \autoref{fig:appendix-curricula-power-2}).

\clearpage

\subsection{Simultaneous training architecture capacity}

We compared a few variations on our model architecture to validate it has the capacity to learn these tasks when trained simultaneously. The loss and AUC curves plotted below provide results from a baseline model, a model with dropout, and the model reported in the paper, which utilizes weight decay but not dropout. 

\begin{figure*}[b!]
\centering
\includegraphics[width=\suppfigwdith\linewidth]{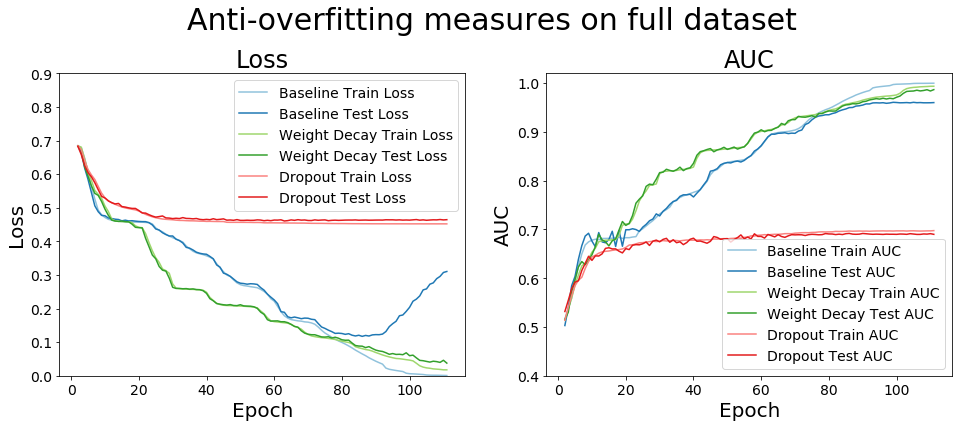}
\caption{Loss and AUC curves for architecture variants when training on all thirty tasks simultaneously.}
\label{fig:appendix-model-capacity}
\end{figure*}

\clearpage

\subsection{Results by dimension}
\begin{figure*}[b!]
\centering
\includegraphics[width=\suppfigwdith\linewidth]{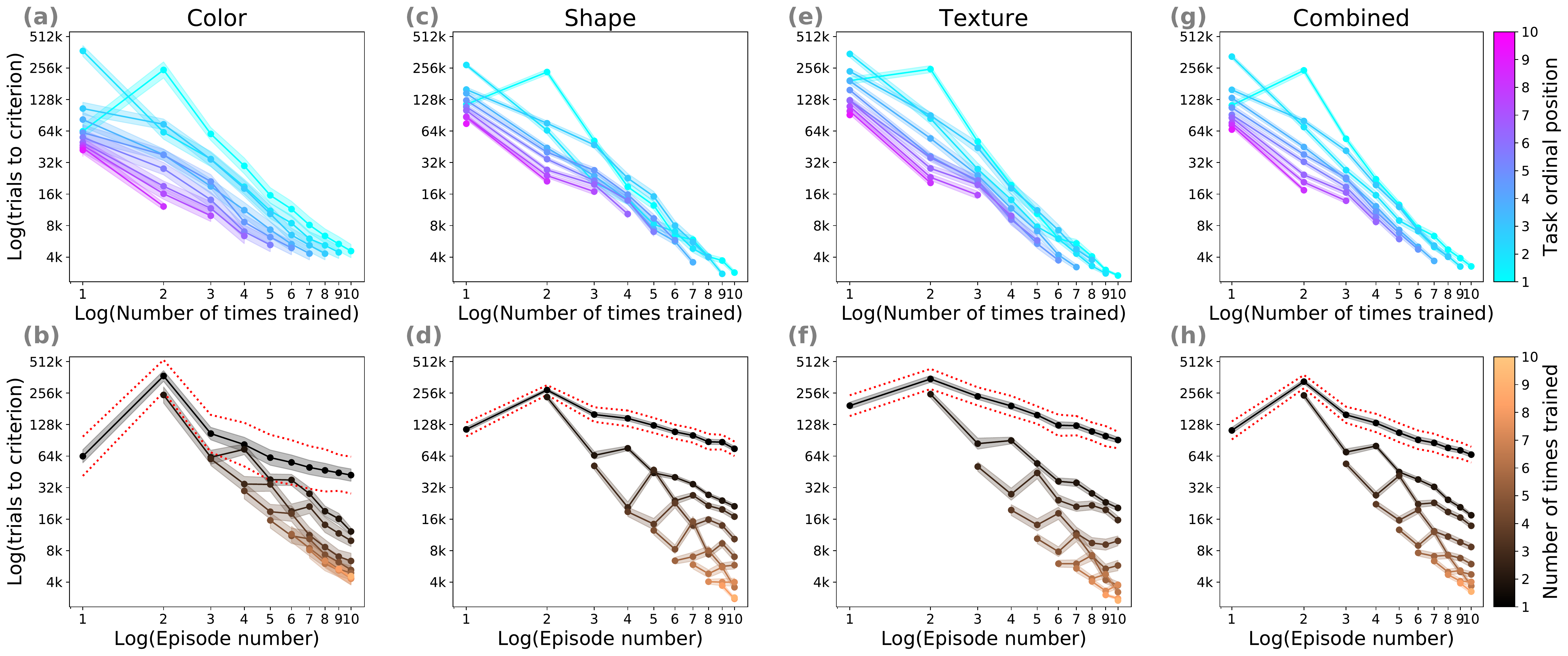}
\caption{\textbf{(a, c, e, g)}: Number of trials required to reach the accuracy criterion (log scale) as a function of the number of times a given task is trained (also log scale). The colored lines indicate task ordinal position (cyan = introduced in episode 1; magenta = introduced in episode 10).
\textbf{(b, d, f, h)}: Number of trials required to reach the accuracy criterion (log scale) as a function of the episode number. The colored lines indicate the number of times a task was retrained on (black = 1 time, copper = 10 times). 
In all panels, the shaded region represents $\pm 1$ standard error of the mean.}
\label{fig:appendix-trials-by-dimension}
\end{figure*}

\begin{figure*}[b!]
\centering
\includegraphics[width=\suppfigwdith\linewidth]{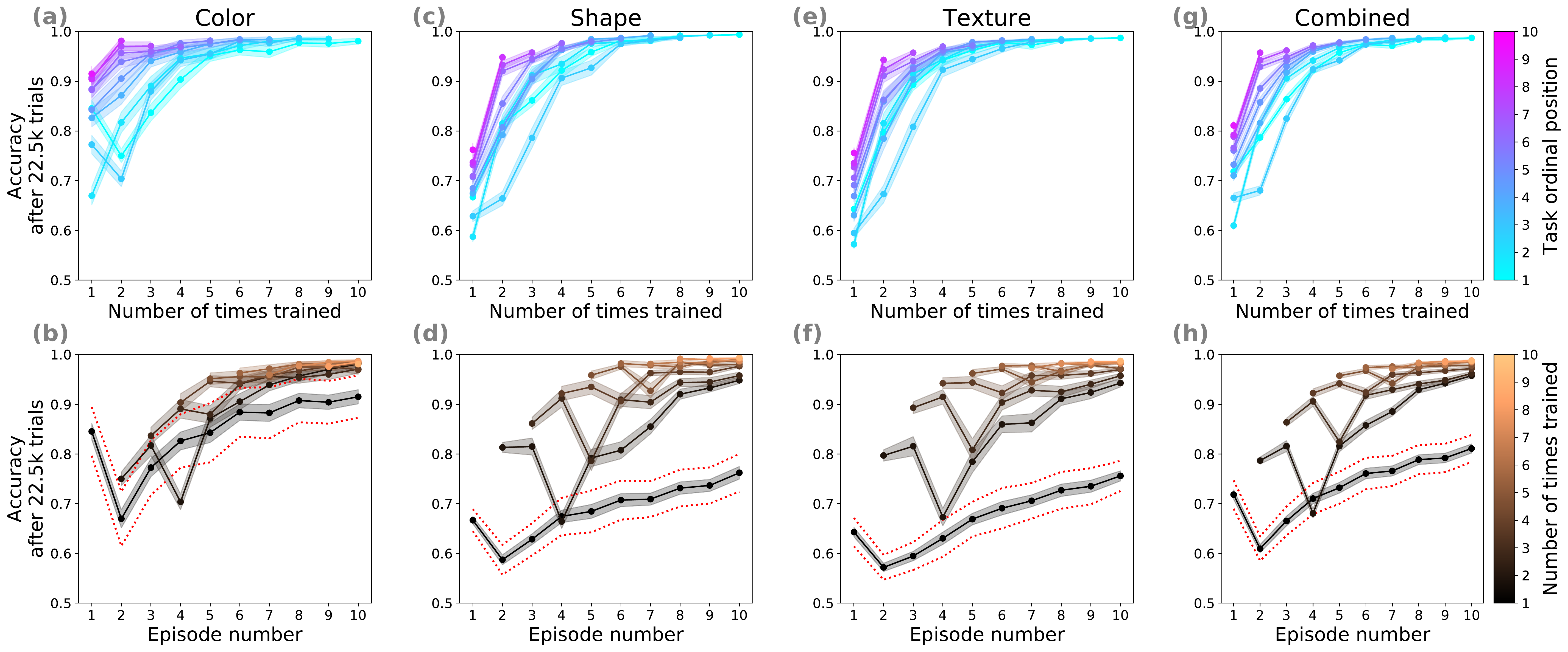}
\caption{\textbf{(a, c, e, g)}: Accuracy after a fixed amount of training (22,500 trials) as a function of the number of times a given task is trained (log scale). The colored lines indicate task ordinal position (cyan = introduced in episode 1; magenta = introduced in episode 10).
\textbf{(b, d, f, h)}: Accuracy after the same fixed amount of training as a function of the episode number. The colored lines indicate the number of times a task was retrained on (black = 1 time, copper = 10 times). 
In all panels, the shaded region represents $\pm 1$ standard error of the mean.}
\label{fig:appendix-accuracy-by-dimension}
\end{figure*}

To justify our collapsing of the results across dimensions, we provide the results broken down for each individual dimension below. \autoref{fig:appendix-trials-by-dimension} depicts the trials required to reach the accuracy criterion, Figure~\ref{fig:appendix-trials-by-dimension}g,h reproducing Figure~3c,d, and the rest of the subfigures offering the results for replications within each dimension. While colors are easier to learn than shapes or textures, simulations in all three dimensions show the same qualitative features. Similarly, \autoref{fig:appendix-accuracy-by-dimension} depicts the accuracy after a fixed small amount of training, with Figure~\ref{fig:appendix-accuracy-by-dimension}g,h reproducing Figure~3e,f. These results provide further evidence for the ease of learning color compared to the other two dimensions, but the qualitative similarity remains striking. 

\clearpage

\subsection{Task-modulated processing at different levels}

\begin{figure*}[b!]
\centering
\includegraphics[width=\suppfigwdith\linewidth]{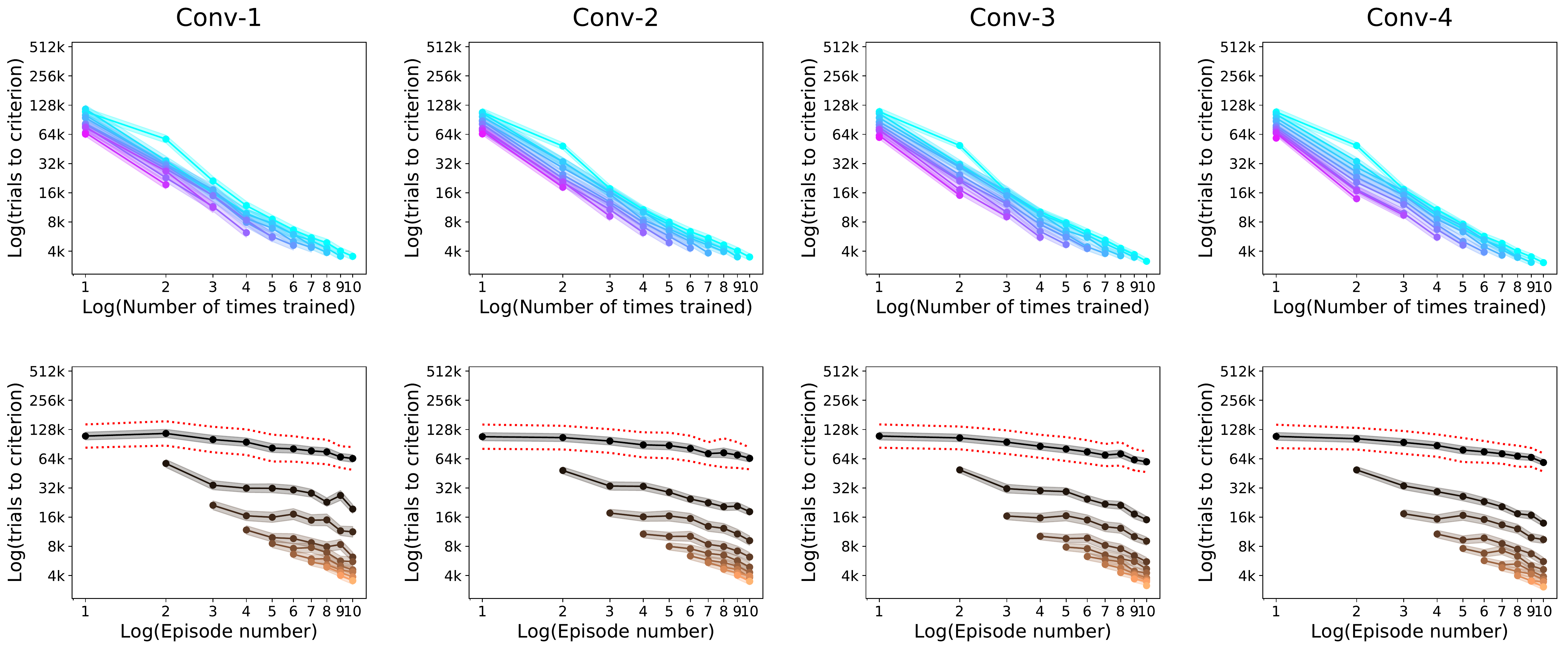}
\caption{\textbf{Top panels:} Number of trials required to reach the accuracy criterion (log scale) as a function of the number of times a given task is trained (also log scale). The colors indicate task ordinal position (the episode in which a task is introduced; cyan = introduced in episode 1; magenta = introduced in episode 10). \textbf{Bottom panels:} Similar to the top panels, but graphed as a function of episode number with the line colors indicating the number of times a task is retrained (black = 1 time, copper = 10 times).}
\label{fig:appendix-task-mod-trials}
\end{figure*}

\begin{figure*}[b!h]
\centering
\includegraphics[width=\suppfigwdith\linewidth]{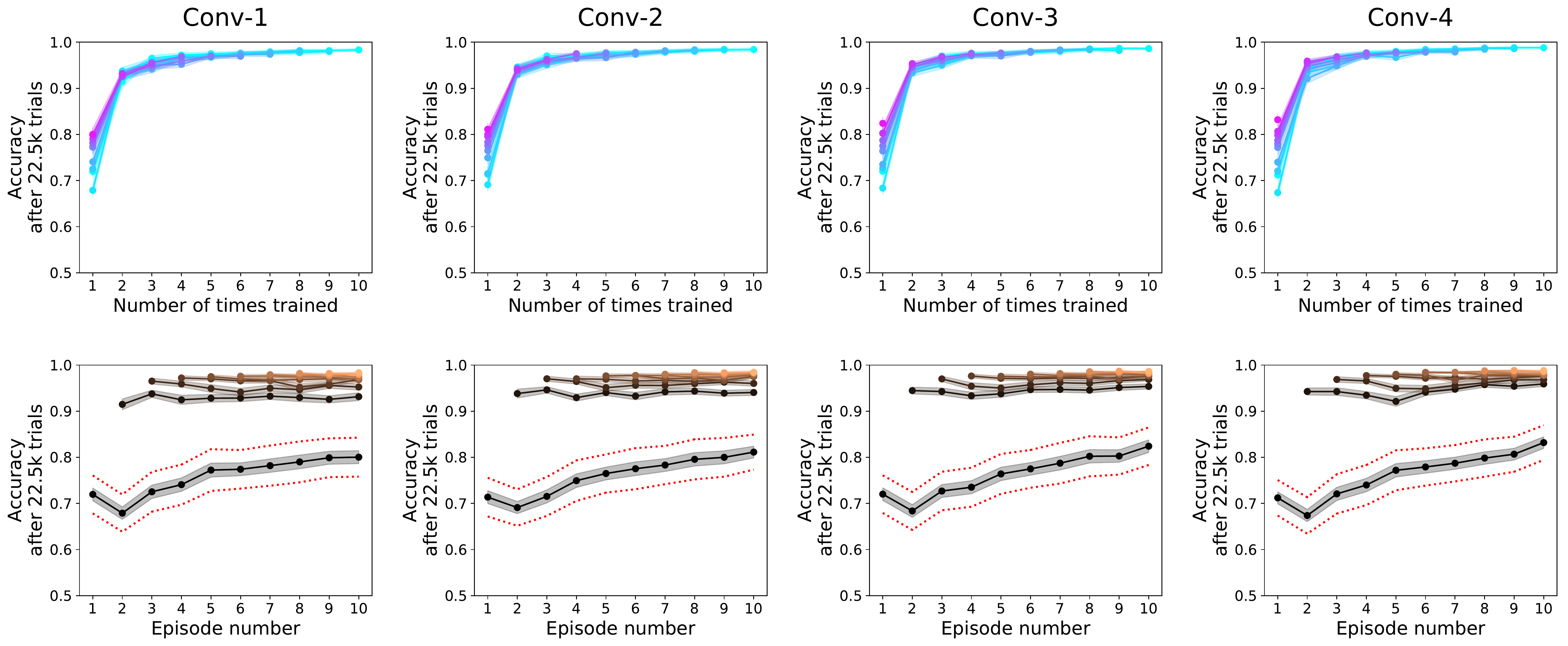}
\caption{\textbf{Top panels:} Hold-out accuracy attained after a fixed amount of training (22.5k trials) of a given task, graphed as a function of number of times a given task is trained. As in \autoref{fig:appendix-task-mod-trials}, the colors indicate task ordinal position (the episode in which a task is introduced; cyan = introduced in episode 1; magenta = introduced in episode 10). \textbf{Bottom panels:} Similar to the top panels, but graphed as a function of episode number with the line colors indicating--as in \autoref{fig:appendix-task-mod-trials}--the number of times a task is retrained (black = 1 time, copper = 10 times).}
\label{fig:appendix-task-mod-accuracy}
\end{figure*}

\begin{figure*}[b!h]
\centering
\includegraphics[width=\suppfigwdith\linewidth]{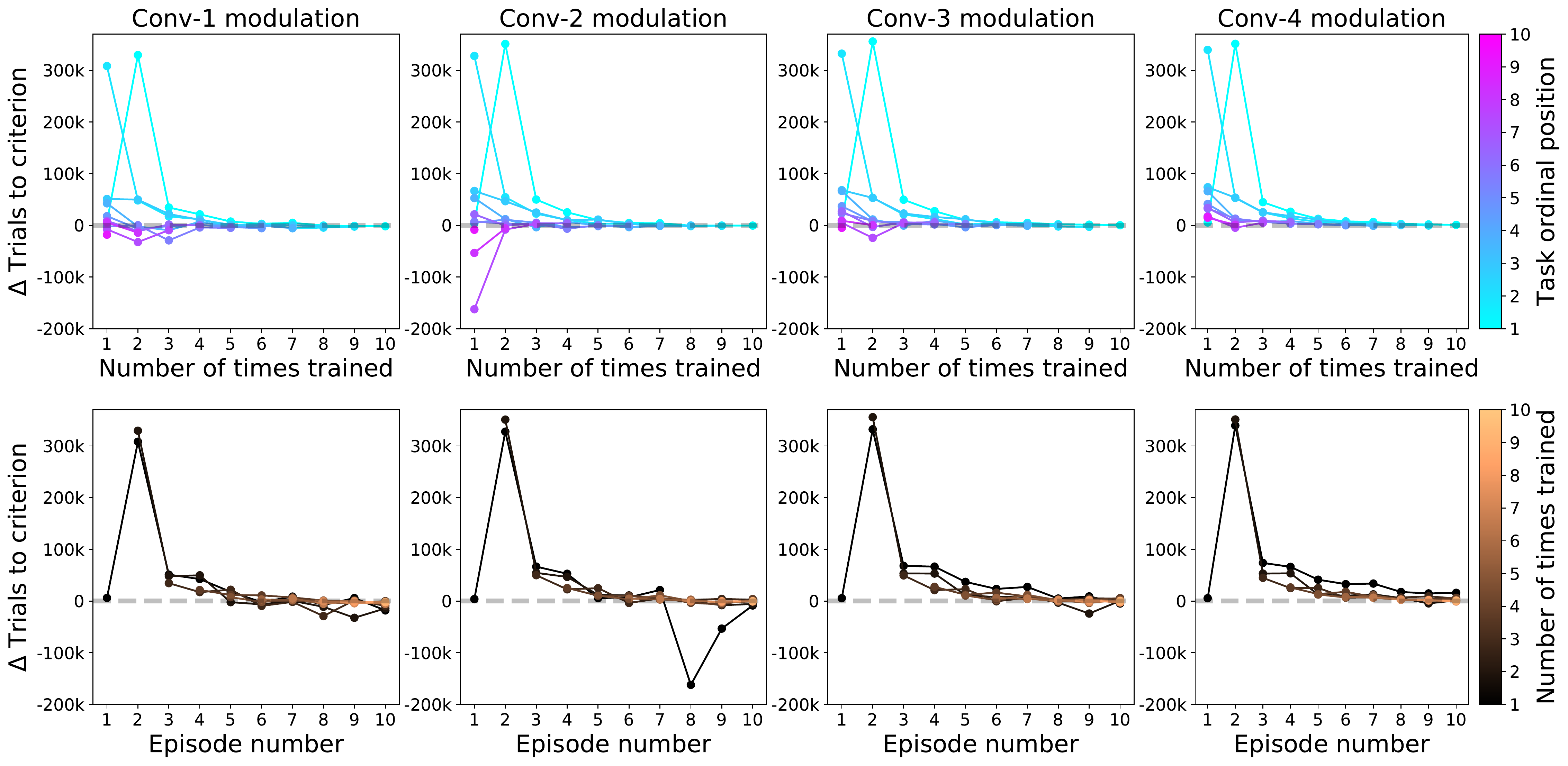}
\caption{\textbf{Top panels:} Increase in number of trials required to reach accuracy criterion  for non-task-modulated versus task modulated architectures as a function of the number of times a given task is trained (also log scale). The colors indicate task ordinal position (the episode in which a task is introduced; cyan = introduced in episode 1; magenta = introduced in episode 10). \textbf{Bottom panels:} Similar to the top panels, but graphed as a function of episode number with the line colors indicating the number of times a task is retrained (black = 1 time, copper = 10 times).}
\label{fig:appendix-task-mod-trials-comparison}
\end{figure*}

All figures reported below are combined over replications in all three dimensions, where for each modulation level we performed thirty simulations in each dimension, yielding ninety simulations in total for each modulation level. 
In \autoref{fig:appendix-task-mod-trials}, we provide the results plotted in Figure~5a-b for task-modulation at each convolutional layer (separately). In \autoref{fig:appendix-task-mod-accuracy}, we provide equivalent plots to Figure~2e-f for the task-modulated models. In \autoref{fig:appendix-task-mod-trials-comparison}, we provide equivalent plots to Figure~5c-d for the task-modulated models. The only anomaly we observe is in \autoref{fig:appendix-task-mod-trials-comparison} for task-modulation at the second convolutional layer, where the eight and ninth tasks appear easier to learn for the first time without task-modulation. Save for this anomaly, we observed remarkably consistent results between the different modulation levels, and hence we reported a single one, rather than expanding about all four. 

\clearpage

\subsection{MAML comparison supplement}
We compared our baseline model to two versions of MAML. Both utilized the training procedure we describe under the `Comparison to MAML' section. 
The first, reported in the middle column below, only utilized this procedure in training, and was tested without the meta-testing step. 
In other words, this model was tested exactly as our baseline model was tested, to see if MAML manages to learn representations that allow it to answer questions on unseen images without further adaptation. 
The second version, which we ended up reporting, also utilizes the micro-episode procedure at test time, making train and test identical. 
The results below demonstrate similar qualitative behavior between our baseline and both versions.
However, as the second version, using the meta-testing procedure, fares better, we opt to report it in the submission.

\begin{figure*}[b!h]
\centering
\includegraphics[width=\suppfigwdith\linewidth]{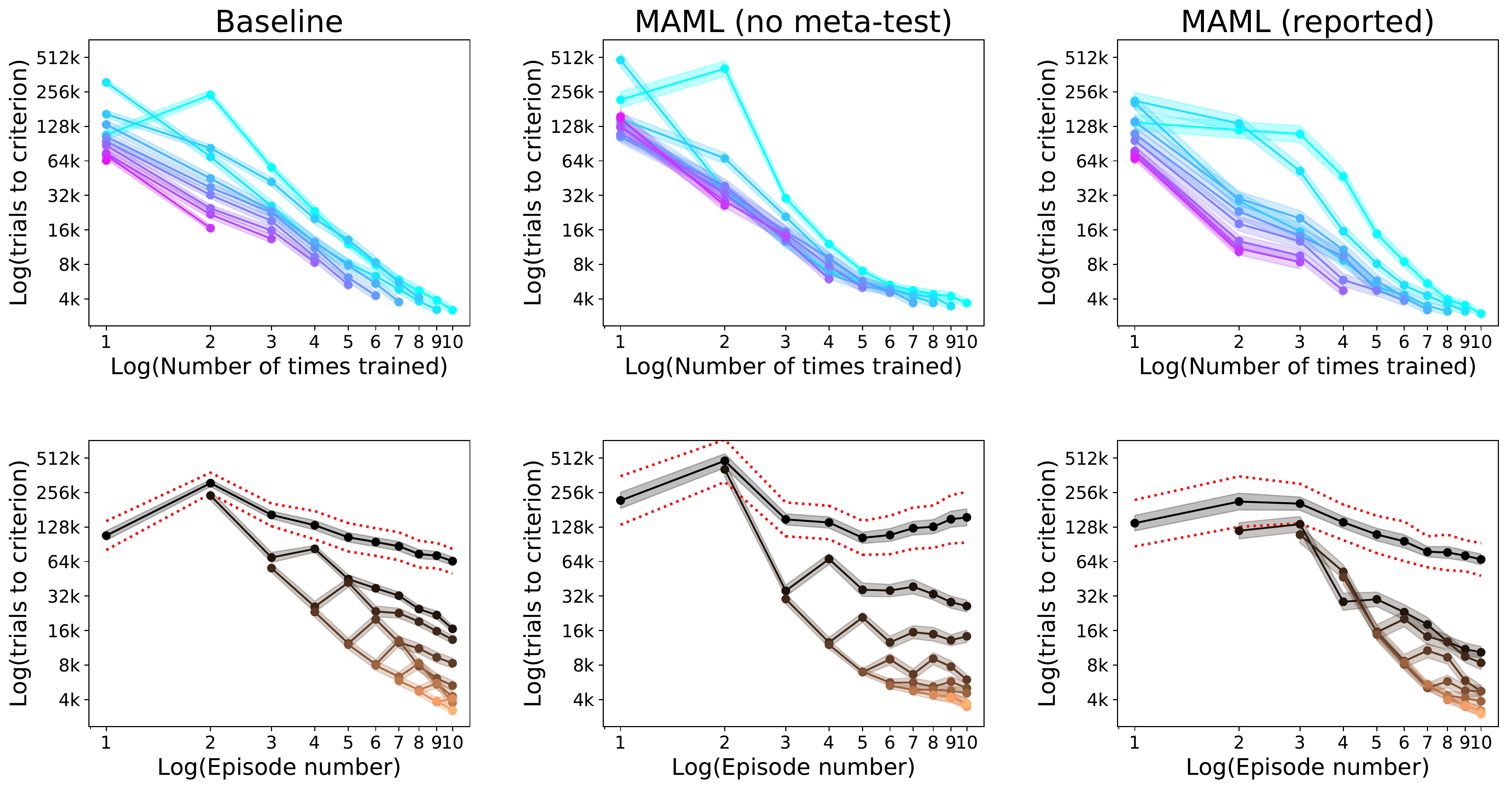}
\caption{MAML comparison. The left column plots result from our baseline condition. The middle column offers results from a version of MAML which did not follow the micro-episode procedure at test that, that is, did not meta-test. The right column, corresponding to the model reported in the submission, follows the micro-episode procedure we describe at both train and test.}
\label{fig:appendix-maml-comparison}
\end{figure*}

\clearpage

\subsection{Comparison to simultaneous learning}
\begin{figure*}[b!h]
\centering
\includegraphics[width=\suppfigwdith\linewidth]{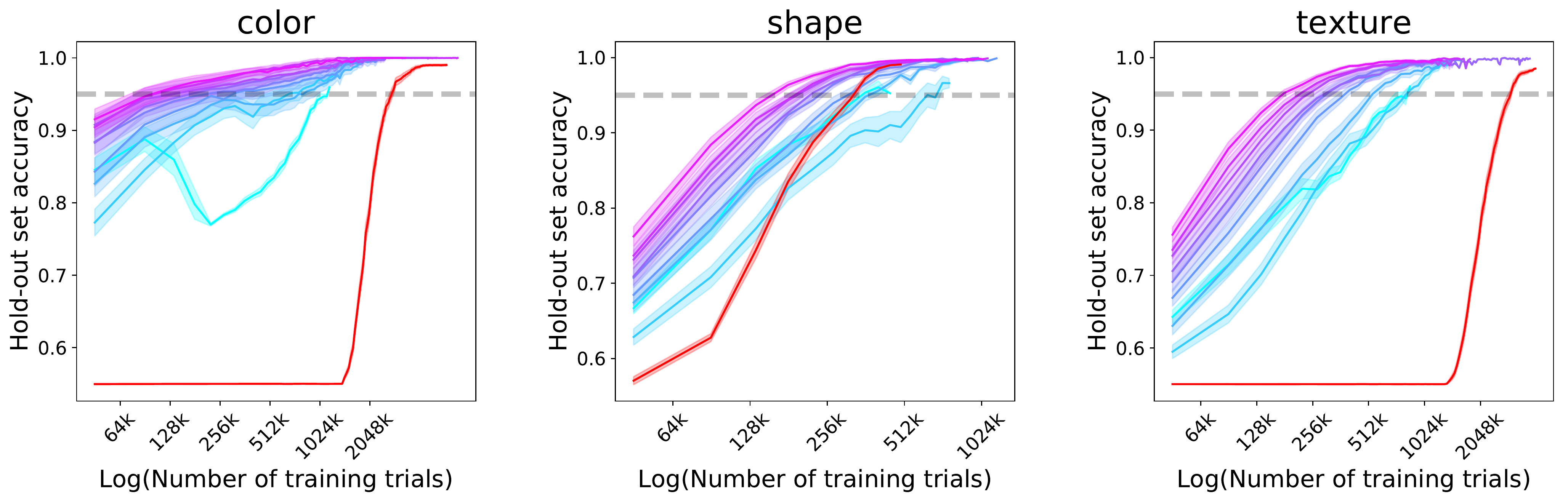}
\caption{Simultaneous vs. sequential training. 
The cyan (first) to magenta (last) colored lines plot the accuracy after some number of training trials for each task the model learned.
The average accuracy over all ten tasks, when learned simultaneously, is plotted in red.
To make the comparisons valid, the simultaneous training is in the number of training trials \emph{for each task}, rather than combined for all tasks.}
%In each panel, accuracy after each number of training trials is plotted in the cyan to magenta colors, while the accuracy trained simultaneously is plotted in red. To make the comparisons valid, the simultaneous training is in the number of training trials \emph{for each task}, rather than combined for all tasks.}
\label{fig:simultaneous}
\end{figure*}

We performed a systematic comparison between our sequential method of training and the standard supervised learning approach of training on all tasks simultaneously. 
We know that sequential training is beneficial to humans--every course covers one topic at a time, rather than throwing the entire textbook and mixing all topics from day one. 
There is also ample evidence for the value of curricular approaches in machine learning, going as far back as \citep{Elman1993}. 
However, curricula in machine learning usually attempt to scaffold tasks from smaller to larger, or easier to harder, following some difficulty gradient.
Our results in \autoref{fig:simultaneous} suggest, surprisingly, that randomly chosen sequential curriculum (that is, random task introduction orderings) can significantly speed up learning in some cases.  
We find, interestingly, that this effect varies by dimension. 
While in the shape condition the simultaneous learning is competitive with sequential training, we find that in both texture and color sequential training proceeds much faster. 
In those cases, the number of training trials required to learn each task when trained sequentially (the cyan-to-magenta curves) is far less than the number of trials required to learn each task when trained simultaneously (the red curve). 
That is, task $n+1$ is learned far faster following tasks 1--$n$ than simultaneously with tasks 1--10. 
The long plateau in the color and texture cases appears to suggest some form of initial representation learning which is made more efficient by learning sequentially, rather than simultaneously. 

%\end{document}

\end{document}